\documentclass[conference]{IEEEtran}
\ifCLASSINFOpdf
\else
\fi
\usepackage[utf8]{inputenc} 
\usepackage[T1]{fontenc}    
\usepackage{hyperref}       
\usepackage{url}            
\usepackage{booktabs}       
\usepackage{amsfonts}       
\usepackage{nicefrac}       
\usepackage{microtype}      
\usepackage{graphicx}
\usepackage{float}
\usepackage{url}
\usepackage{amsmath}
\usepackage{comment}
\usepackage{diagbox}
\usepackage{makecell}
\usepackage{diagbox}

\usepackage{amsmath,amsfonts,bm}

\def\eqref#1{equation~\ref{#1}}

\def\1{\bm{1}}

\def\vy{{\bm{y}}}

\def\mM{{\bm{M}}}

\def\mV{{\bm{V}}}
\def\mW{{\bm{W}}}
\def\mX{{\bm{X}}}

\DeclareMathAlphabet{\mathsfit}{\encodingdefault}{\sfdefault}{m}{sl}
\SetMathAlphabet{\mathsfit}{bold}{\encodingdefault}{\sfdefault}{bx}{n}

\usepackage{color}
\usepackage{amsmath}
\usepackage{sidecap}
 \usepackage[numbers]{natbib}
\usepackage{algorithm}
\usepackage{algpseudocode}
\usepackage{float}
\usepackage{flushend} 

\begin{document}
%
\title{Beneficial Perturbation Network for designing general adaptive artificial intelligence systems}


%
\author{\IEEEauthorblockN{Shixian Wen\IEEEauthorrefmark{1},
Amanda Rios\IEEEauthorrefmark{2}\IEEEauthorrefmark{4},
Yunhao Ge\IEEEauthorrefmark{1} \IEEEauthorrefmark{4} and  
Laurent Itti\IEEEauthorrefmark{1} \IEEEauthorrefmark{2} \IEEEauthorrefmark{3}}
\IEEEauthorblockA{\IEEEauthorrefmark{1}Department of Computer Science\\
University of Southern California,
Los Angeles, California 90089\\ Email: shixianw@usc.edu}
\IEEEauthorblockA{\IEEEauthorrefmark{2} Neuroscience Graduate Program,
University of Southern California}
\IEEEauthorblockA{\IEEEauthorrefmark{3}Department of psychology,
University of Southern California}
\IEEEauthorblockA{\IEEEauthorrefmark{4}Contributed equally as the second authors}}


\maketitle

\begin{abstract}
The human brain is the gold standard of adaptive learning. It not only can learn and benefit from experience, but also can adapt to new situations. In contrast, deep neural networks only learn one sophisticated but fixed mapping from inputs to outputs. This limits their applicability to more dynamic situations, where the input to output mapping may change with different contexts. A salient example is continual learning - learning new independent tasks sequentially without forgetting previous tasks. Continual learning of multiple tasks in artificial neural networks using gradient descent leads to catastrophic forgetting, whereby a previously learned mapping of an old task is erased when learning new mappings for new tasks. Here, we propose a new biologically plausible type of deep neural network with extra, out-of-network, task-dependent biasing units to accommodate these dynamic situations. This allows, for the first time, a single network to learn potentially unlimited parallel input to output mappings, and to switch on the fly between them at runtime. Biasing units are programmed by leveraging beneficial perturbations (opposite to well-known adversarial perturbations) for each task.  Beneficial perturbations for a given task bias the network toward that task, essentially switching the network into a different mode to process that task. This largely eliminates catastrophic interference between tasks. Our approach is memory-efficient and parameter-efficient, can accommodate many tasks, and achieves state-of-the-art performance across different tasks and domains.

\end{abstract}


%
\IEEEpeerreviewmaketitle

\section{Introduction}

\begin{figure}[hb!]
	\begin{center}
		\includegraphics[width=0.8\linewidth]{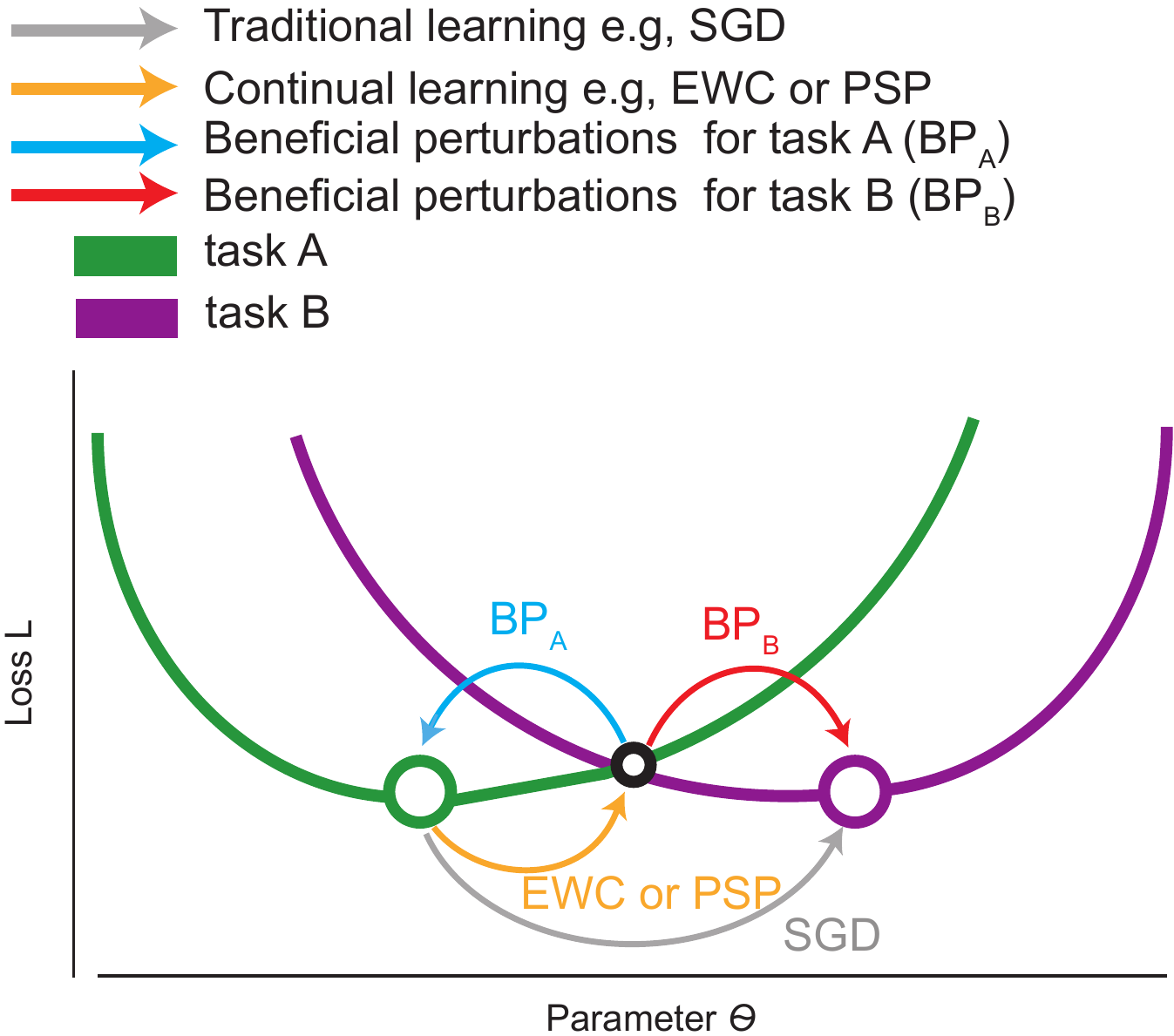}
		\caption{ {With BPN, one can switch at runtime the network parameters that are global optimal for each task. Training trajectories are illustrated in loss and parameter space. The green curve shows loss as a function of network parameters for a first task A, with optimal parameters shown by the green circle. The purple curve and circle correspond to a second task B. Training first task A then task B with stochastic gradient descend (SGD, without any constraints on parameters, gray) leads to optimal parameters for task B (purple circle), but those are destructive for task A. When, instead, learning task B using EWC or PSP (have some constraints on parameters, yellow), the solution is a compromise that can be sub-optimal for both tasks (black circle). Beneficial perturbations (blue curve for task A, red curve for task B) push the representation learned by EWC or PSP back to their task-optimal states.}}
		\label{fig:concept_in_loss_space}
	\end{center}
\end{figure}

 {The human brain is the benchmark of adaptive learning. While interacting with new environments that are not fully known to an individual, it is able to quickly learn and adapt its behavior to achieve goals as well as possible, in a wide range of environments, situations, tasks, and problems. In contrast, deep neural networks only learn one sophisticated but fixed mapping between inputs and outputs, thereby limiting their application in more complex and dynamic situations in which the mapping rules are not kept the same but change according to different tasks or contexts. One of the failed situations is continual learning - learning new independent tasks sequentially without forgetting previous tasks. In the domain of image classification, for example, each task may consist of learning to recognize a small set of new objects. A standard neural network only learns a fixed mapping rule between inputs and outputs after training on each task. Training the same neural network on a new task would destroy the learned fixed mapping of an old task. Thus, current deep learning models based on stochastic gradient descent suffer from so-called "catastrophic forgetting" \citep{mccloskey1989catastrophic,french1994dynamically,sloman1992episodic}, in that they forget all previous tasks after training each new one.}

 {Here, we propose a new biological plausible (Discussion) method~--- Beneficial Perturbation Network (BPN)~--- to accommodate these dynamic situations. The key new idea is to allow one neural network to learn potentially \textit{unlimited} task-dependent mappings and to switch between them at runtime. To achieve this, we first leverage existing lifelong learning methods to reduce interference between successive tasks (Elastic Weight Consolidation, EWC \citep{kirkpatrick2017overcoming}, or parameter superposition, PSP \cite{cheung2019superposition}). We then add out-of-network, task-dependent bias units, to provide per-task correction for any remaining parameter drifts due to the learning of a sequences of tasks. We compute the most beneficial biases~---~beneficial perturbations~---~for each task in a manner inspired by recent work on adversarial examples. The central difference is that, instead of adding adversarial perturbations that can force the network into misclassification, beneficial perturbations can push the drifted representations of old tasks back to their initial task-optimal working states (Fig.~\ref{fig:concept_in_loss_space}).}

\begin{figure*}[htb]
	\begin{center}
		\includegraphics[height = 15cm]{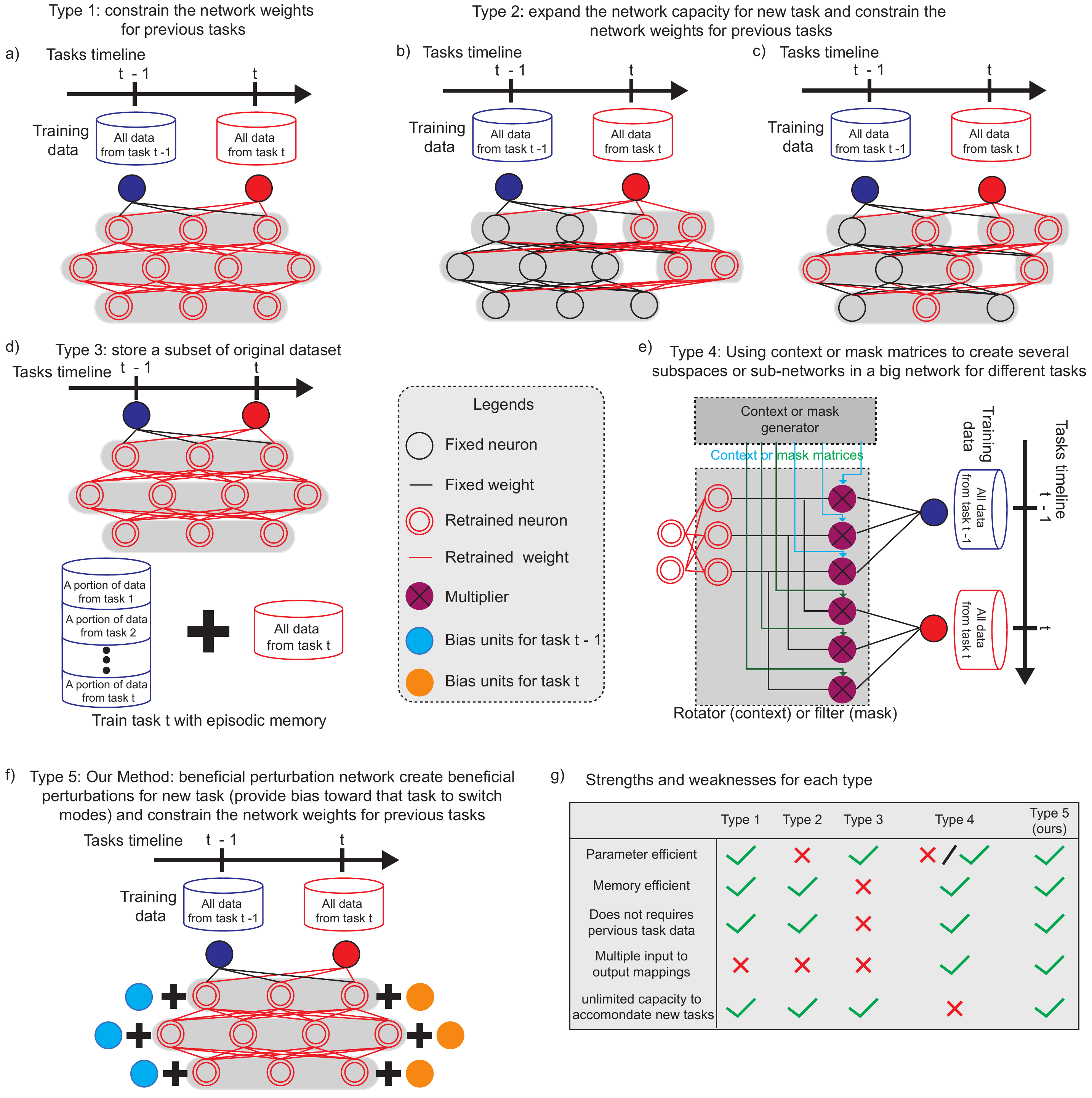}
		\caption{{\bf Concept:} Type 1 - constrain the network weights while training the new task: (a) Retraining models such as elastic weight consolidation \citep{kirkpatrick2017overcoming}: retrains the entire network learned on previous tasks while using a regularizer to prevent drastic changes in the original model. Type 2 - expanding and retraining methods (b-c); (b) Expanding models such as progressive neural networks  \citep{rusu2016progressive} expand the network for new task \textit{$t$} without any modifications to the network weights for previous tasks. (c) Expanding model with partial retraining such as dynamically expandable networks \citep{yoon2018lifelong} expand the network for new task t with partial retraining on the network weights for previous tasks. Type 3 - episodic memory methods (d): Methods such as Gradient Episodic Memory \citep{lopez2017gradient}  store a subset of the original dataset from previous tasks into the episodic memory and replays them with new data during the training of new tasks. Type 4 - Partition network (e): these use context or mask matrices to partition the core network into several sub-networks for different tasks \citep{cheung2019superposition,zeng2019continual,mallya2018piggyback,du2019single,yoon2019oracle}. Type 5 - beneficial perturbation methods (f):  Beneficial perturbation networks create beneficial perturbations which are stored in bias units for each task. Beneficial perturbations bias the network toward that task and thus allow the network to switch into different modes to process different independent tasks. It retrains the normal weights learned from previous tasks using elastic weight consolidation \citep{kirkpatrick2017overcoming} or parameter superposition \citep{cheung2019superposition}. (g) Strengths and weaknesses for each type of method.}
		\label{fig:concept}
	\end{center}
\end{figure*}


There are three major benefits of BPN: {\bf{1)}} BPN is memory and parameter efficient: to demonstrate it, we validate our BPN for continual learning on incremental tasks. We test it on multiple public datasets (incremental MNIST \citep{lecun1998gradient}, incremental CIFAR-10  and incremental CIFAR-100 \citep{krizhevsky2009learning}), on which it achieves better performance than the state-of-the-art. For each task, by adding bias units that store beneficial perturbations to every layer of a 5-layer fully connected network, we only introduce a 0.3\% increase in parameters, compared to a 100\% parameter increase for models that train a separate network, and 11.9\% - 60.3\% for dynamically expandable networks \citep{yoon2018lifelong}. Our model does not need any episodic memory to store data from the previous tasks and does not need to replay them during the training of new tasks, compared to episodic memory methods \citep{rebuffi2017icarl,lopez2017gradient,rannen2017encoder,rios2018closed}. Our model does not need large context matrices, compared to partition methods \citep{cheung2019superposition,zeng2019continual,yoon2019oracle,mallya2018piggyback,du2019single,farajtabar2020orthogonal,srivastava2013compete,masse2018alleviating}.  {\bf{2)}} BPN achieves state-of-the-art performance across different datasets and domains: to demonstrate it, we consider a sequence of eight unrelated object recognition datasets (Experiments). After training on the eight complex datasets sequentially, the average test accuracy of BPN is better than the state-of-the-art. {\bf{3)}} BPN has capacity to accommodate a large number of tasks: to demonstrate it, we test a sequence of 100 permuted MNIST tasks (Experiments).  A variant of BPN that uses PSP to constrain the normal network achieves 30.14\% better performance than the second best, the original PSP \citep{cheung2019superposition}, a partition method which performs well in incremental tasks and eight object recognition tasks. Thus, BPN has a promising future to solve continual learning compared to the other types of methods.

 {To lay out the foundation of our approach we start by introducing the following key concepts: Sec.~\ref{sectypes}: Different types of methods for enabling lifelong learning; Sec.~\ref{adp}: Adversarial directions and perturbations; Sec.~\ref{bdp}: Beneficial directions and  perturbations, and the effects of beneficial perturbations in sequential learning scenarios; Sec.~\ref{bpn}: Structure and updating rules for BPN.}

 {We then present experiments (Sec.~\ref{experiments}), results (Sec.~\ref{results}) and discussion (Sec.~\ref{discussion}).}

\section{Types of methods for enabling lifelong learning}
\label{sectypes}

Four major types of methods have been proposed to alleviate catastrophic forgetting. Type 1: constrain the network weights to preserve performance on old tasks while training the new task \citep{kirkpatrick2017overcoming,lee2017overcoming,
aljundi2018memory} (Fig.~\ref{fig:concept}a); A famous example of type 1 methods is EWC \citep{kirkpatrick2017overcoming}. EWC constrains certain parameters based on how important they are to previously seen tasks. {The importance is calculated from their task-specific Fisher information matrix. However, solely relying on constraining the parameters of the core network eventually exhausts the core network's capacity to accommodate new tasks. After learning many tasks, EWC cannot learn anymore because the parameters become too constrained (see Results).} Type 2: dynamic network expansion \citep{li2017learning,lee2017overcoming,rusu2016progressive,yoon2018lifelong} creates new capacity for the new task, which can often be combined with constrained network weights for previous tasks (Fig.~\ref{fig:concept}b-c);  {However, this type is not scalable because it is not parameter efficient (e.g., 11.9\% - 60.3\% additional parameters per task for dynamically expandable networks \citep{yoon2018lifelong})}.  Type 3: using an episodic memory \citep{rebuffi2017icarl,lopez2017gradient,rannen2017encoder} to store a subset of the original dataset from previous tasks, then rehearsing it while learning new tasks to maintain accuracy on the old tasks (Fig.~\ref{fig:concept}d).  { However, this type is not scalable because it is neither memory nor parameter efficient.} All three approaches attempt to shift the network's single fixed mapping initially obtained by learning the first task to a new one that satisfies both old and new tasks. They create a new, but still fixed mapping from inputs to outputs across all tasks so far, combined. Type 4: Partition Network: using task-dependent context \citep{cheung2019superposition,zeng2019continual,yoon2019oracle,masse2018alleviating} or mask matrices \citep{mallya2018piggyback,du2019single,du2019single,farajtabar2020orthogonal,srivastava2013compete} to partition the original network into several small sub-networks (Fig.~\ref{fig:concept}e, flow chart - Fig.~\ref{fig:Flow_charts}a). Zeng {\em et al.} \cite{zeng2019continual} used context matrices to partition the network into independent subspaces spanned by rows in the weight matrices to avoid interference between tasks. However, context matrices introduce as many additional parameters as training a separate neural network for each new task (additional 100\% parameters per task). To reduce parameter costs, Cheung {\em et al.} proposed binary context matrices \citep{cheung2019superposition}, further restricted to diagonal matrices with -1 and 1 values. The restricted context matrices \citep{zeng2019continual} (1 and -1 values) behave similarly to mask matrices \citep{mallya2018piggyback} (0 and 1 values) that split the core network into several sub-networks for different tasks. With too many tasks, the core network would eventually run out of capacity to accommodate any new task, because there is no vacant route or subspace left. Although type 4 methods create multiple input to output mappings for different tasks, many of these methods are too expensive in terms of parameters, and none of them has enough capacity to accommodate numerous tasks because methods such as PSP run out of unrealized capacity of the core network. 

In marked contrast to the above artificial neural network methods, here, we propose a fundamentally new fifth type (Fig.~\ref{fig:concept}f, flow chart - Fig.~\ref{fig:Flow_charts} b): We add out-of-network, task-dependent bias units to neural network. Bias units enable a neural network to switch into different modes to process different independent tasks through beneficial perturbations (the memory storage cost of these new bias units is actually lower than the cost of adding a new mask or context matrix). With only an additional 0.3\% of parameters per mode \footnote{ {Check supplementary discussion for more information about additional parameter costs}}, this structure allows BPN to learn potentially unlimited task-dependent mappings from inputs to outputs for different tasks. The strengths and weaknesses of each type are in Fig.~\ref{fig:concept}g.

\begin{figure*}[htb]
	\begin{center}
		\includegraphics[width=0.9\linewidth]{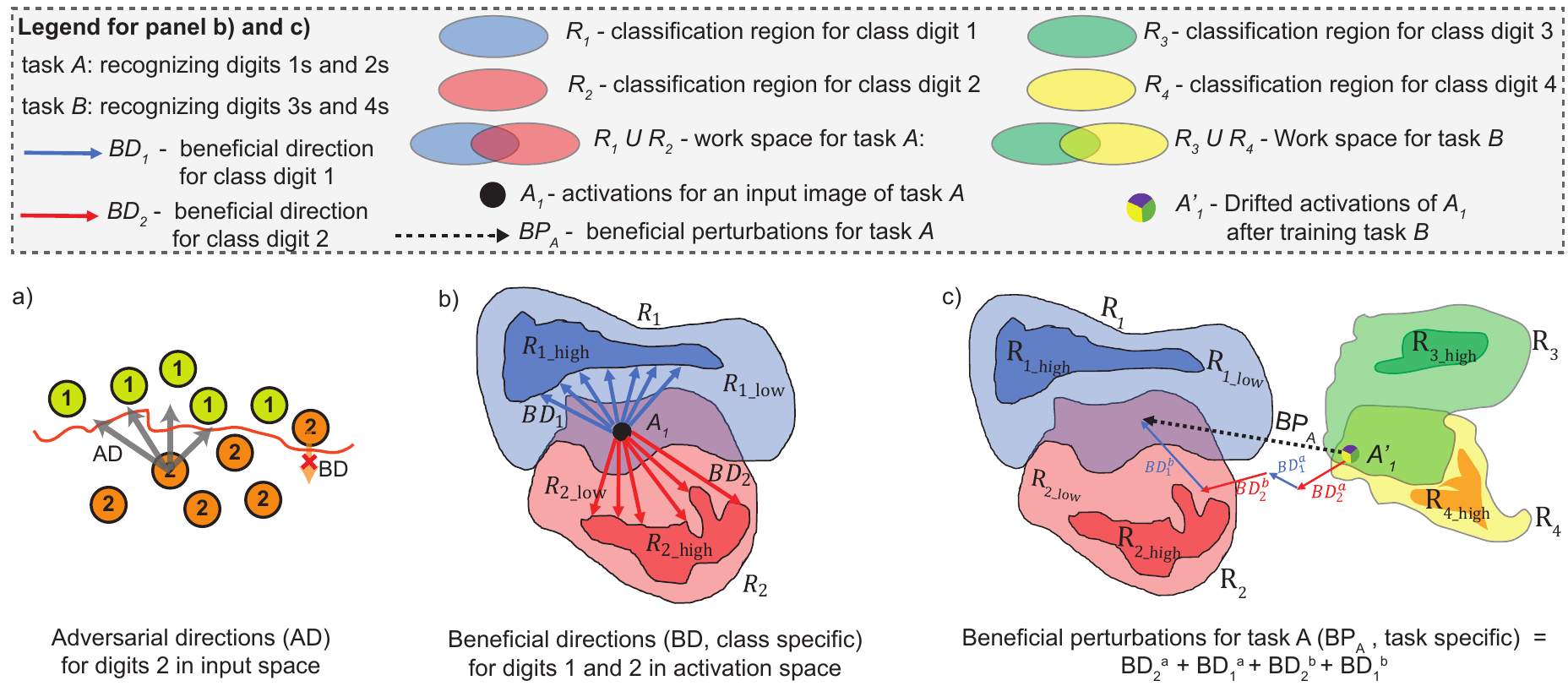}
		\caption{Defining adversarial perturbations in input space vs. beneficial perturbations in activation space.  We consider two digits recognition tasks; Task A (recognizing 1s and 2s) and Task B (recognizing 3s and 4s). (a) {\bf Adversarial directions (AD)}. Adding adversarial perturbations (calculated from digits 1) to input digits 2 can be viewed as adding an adversarial direction vector (gray arrow) to the clear input image of digit 2 in the input space. Thus, the network misclassifies the clear input image of digit 2 as digit 1.  Beneficial directions are not operated as adding beneficial perturbations to the clear input image of digit 2 in the input space to assist the correct classification (orange arrow). (b) {\bf Beneficial directions (class specific) for each class of task A .} $R_1$ ($R_2$) is the classification region (region of constant estimated label) of digit 1 (digit 2) from the MNIST dataset. Subregion $R_{1\_high}$ ($R_{1\_low}$) is the high (low)  confidence classification region of digit 1, and likewise for $R_{2\_high}$ ($R_{2\_low}$) for digit 2. The point $A_1$ is the activations of normal neurons of each layer from an input image of task A. It lies in the intersection of $R_{1\_low}$ and $R_{2\_low}$. $BD_1$ ($BD_2$) are beneficial directions  for class digit 1 (digit 2). $A_1 + BD_1$, blue arrows,($A_+BD_2$, red arrows) pushes the activation $A_1$ across the decision boundary of $R_2$ ($R_1$) and towards $R_{1\_high}$ ($R_{2\_high}$). Thus, the network classifies $A_1 + BD_1$ ($A_1 + BD_2$) as digit 1 (digit 2) with high confidence. (c)  {{\bf  After training task B, beneficial perturbations (task specific) for task A push the drifted representation of inputs from task A back to its initial optimal working region of task A.}} $R_3$ ($R_4$ ) is the classification region (region of constant estimated label) of digit 3 (digit 4) from the MNIST dataset. $BD_1$ ($BD_2$) is a beneficial direction for digit 1 (digit 2). During the training of task A, the network has been trained on two images from digit 1 ($1^a$ and $1^b$) and two images from digit 2 ($2^a$ and $2^b$). Thus, the beneficial perturbations for task A are the vector ($BD_2^{a} + BD_1^{a} + BD_2^{b} + BD_1^{b}$). After training task B, with gradient descent, point $A_1$ in b) is drifted to the $A'_1$ which lies inside of the classification regions of task B ($R_2$ or $R_3$). The drifted point $A'_1$ alone cannot be correctly classified as digit 1 or 2 because it lies outside of the classification region of task A ($R_1$ or $R_2$). At test time, adding beneficial perturbations for task A to the activations of $A'_1$, can drag it back the correct classification regions for task A (intersection of $R_1$ and $R_2$). Thus, it biases the network's outputs toward the correct classification region and push task representations back to  { their initial task-optimal working region.} 
		}\label{fig:beneficial_perturbations}
	\end{center}
\end{figure*}

\section{Adversarial directions and perturbations}
\label{adp}

 {Three spaces of a neural network are important for this and the following sections: The {\em input space} is the space of input data (e.g., pixels of an image); the {\em parameter space} is the space of all the weights and biases of the network; the {\em activation space} is the space of all outputs of all neurons in all layers in the network.}

By adding a carefully computed “noise” (adversarial perturbations) to the input space of a picture, without changing the neural network, one can force the network into misclassification. The noise is usually computed by backpropagating the gradient in a so-called “adversarial direction” such as by using the fast gradient sign method (FGSD) \citep{tramer2017space}. For example,  consider a task of recognizing handwritten digits "1" versus "2". Adversarial perturbations aimed at misclassifying an image of digit 2 as digit 1 may be obtained by backpropagating from the class digit 1 to the input space, following any of the available adversarial directions.  In Fig.~\ref{fig:beneficial_perturbations}a, adding adversarial perturbations to the input image can be viewed as adding an adversarial direction vector  (gray arrows $AD$) to the clear (non-perturbated) input image of digit 2. The resulting vector crosses the decision boundary. Thus, adversarial perturbations can force the neural network into misclassification, here from digit 2 to digit 1. Because the dimensionality of adversarial directions is around 25 for MNIST \citep{tramer2017space}, when we project them into a 2D space, we use the fan-shaped gray arrows to depict those dimensions. 

\section{Beneficial directions and perturbations, \& The effects of beneficial perturbations in multitask sequential learning scenario}
\label{bdp}
{In this section, we first introduce the definition of beneficial directions and beneficial perturbations. Then, we explain why beneficial perturbations can help a network recover from a parameter drifting of old tasks after learning new tasks and can push task representations back to their initial task-optimal working region.}

\begin{figure*}[htb]
	\begin{center}
		\includegraphics[width=0.8\linewidth]{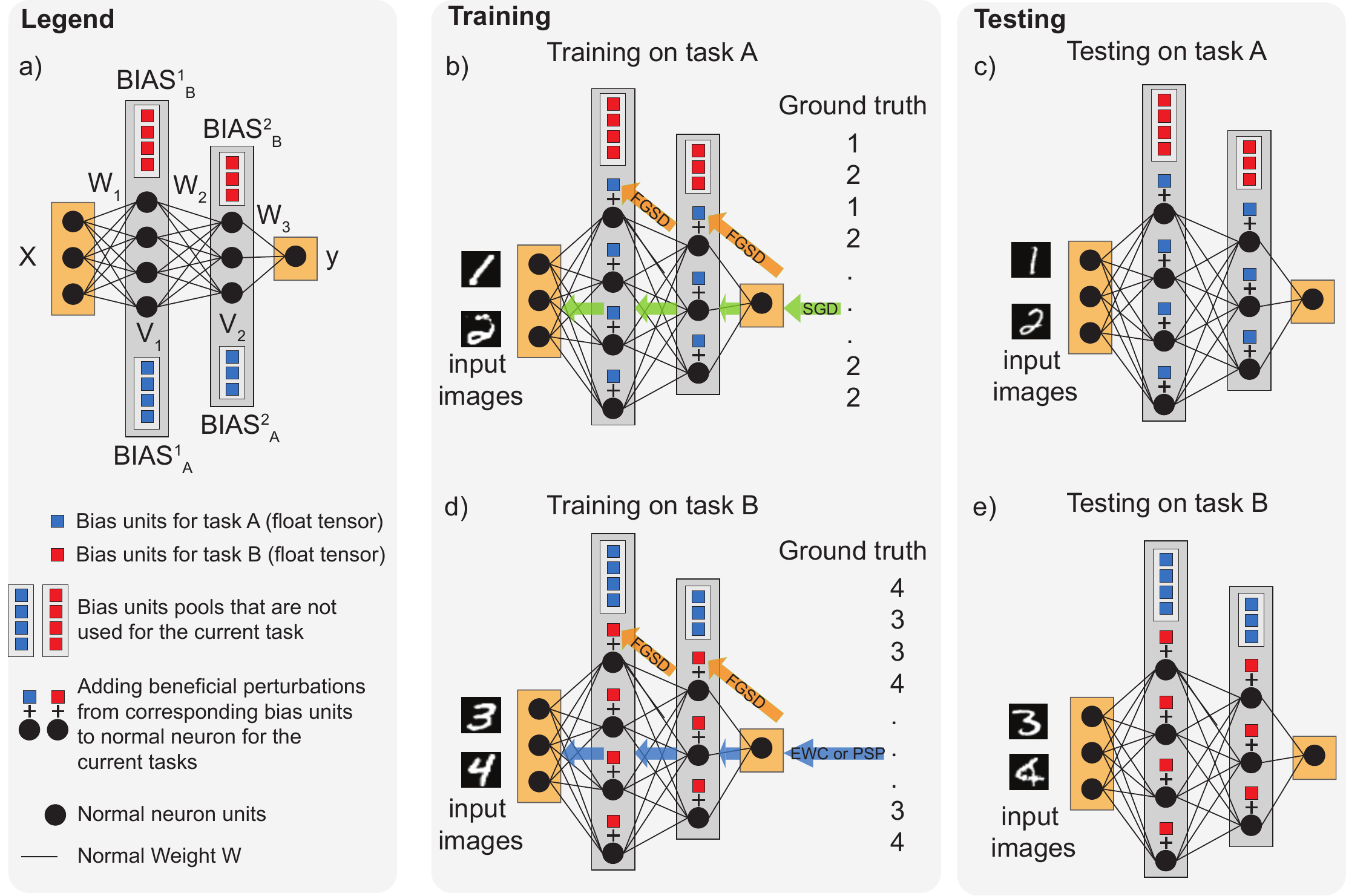}
		\caption{  {\bf{ Beneficial perturbation network (BD + EWC or BD + PSP variant) with two tasks.}} (a) Structure of beneficial perturbation network.  (b) Train on task A. Backpropagating through the network to bias units for tasks A in beneficial direction (FGSD) using input's own correct class (digits label 1 and 2), normal weights (gradient descent). (c) Test on task A. Feed the input images to the network. Activating bias units for task A and adding the stored beneficial perturbations to the activations. The beneficial perturbations bias the network to mode on classifying digits 1, 2 task. (d) Train on task B. Backpropagating through the network to bias units for tasks B in beneficial direction (FGSD) using input's own correct class (digits label 3 and 4), normal weights (constrained by EWC or PSP). (e) Test on task B.  Feed the input images to the network. Activating bias units for task B and adding the stored beneficial perturbations to the activations. The beneficial perturbations bias the network to mode on classifying digits 3, 4 task.}
		\label{fig:explanation_structure}
	\end{center}
\end{figure*}

We consider two incremental digits recognition tasks; Task A (recognizing 1s and 2s) and Task B (recognizing 3s and 4s). Attack and defense researchers usually view adversarial examples as a curse of neural networks, but we view it as a gift to solve continual learning. Instead of adding input "noise" (adversarial perturbations) to the {\em input space} calculated from other classes to force the network into misclassification,  {we add "noise" to the {\em activation space}, using {\em beneficial perturbations} stored in bias units added to the {\em parameter space} (Supplementary Fig.~\ref{fig:Flow_charts}b) calculated by the input's own correct class to assist in correct classification.} To understand beneficial perturbations, we first explain beneficial directions. Beneficial directions are vectors that point toward the direction of high confidence classification region for each class (Fig.~\ref{fig:beneficial_perturbations}b); ${BD}_1$ (${BD}_2$) are the beneficial directions that point to the high confidence classification region of digit 1 (digit 2). The point $A_1$ represents the activation of the normal neurons of each layer generated from an input image of task A. $A_1 + {BD}_1$ ($A_1 + {BD}_2$) pushes the activation $A_1$ across the decision boundary of $R_2$ ($R_1$) and toward $R_{1\_high}$ ($R_{2\_high}$). Thus, the network would classify the $A_1 + {BD}_1$ ($A_1 + {BD}_2$) as digit 1 (2) with high confidence.
To overcome catastrophic forgetting, we create some beneficial perturbations for each task and store them in task-dependent bias units (Fig.~\ref{fig:explanation_structure}, Supplementary Fig.~\ref{fig:Flow_charts}b). Beneficial perturbations allow a neural network to operate in different modes by biasing the network toward that particular task, even though the shared normal weights become contaminated by other tasks. The beneficial perturbations for each task are created by aggregating the beneficial direction vectors sequentially for each class through mini-batch backpropagation. For example, during the training of task A, the network has been trained on two images from digit 1 ($1^a$ and $1^b$) and two images from digit 2 ($2^a$ and $2^b$). The beneficial perturbations for task A are the summation of the beneficial directions calculated from each image ($BD_2^{a} + BD_1^{a} + BD_2^{b} + BD_1^{b}$ in Fig. ~\ref{fig:beneficial_perturbations}c,  { $BD_i^j$ is the beneficial direction for sample $j$ in class $i$}). During the training of task B, with gradient descent, the point $A_1$ (Fig.~\ref{fig:beneficial_perturbations}b) is drifted to $A'_1$ which lies inside the classification regions for task B ($R_3\bigcup R_4$). The drifted $A'_1$ alone cannot be classified as digit 1 or 2 since it lies outside of the classification regions of task A ($R_1 \bigcup R_2$).  However, during testing of task A, after training task B, adding beneficial perturbations for task A to the drifted activation ($A'_1$) drags it back to the correct classification regions for task A ( $R_1$ $\bigcup$ $R_2$ in Fig.~\ref{fig:beneficial_perturbations}c). Thus, beneficial perturbations bias the neural network toward that task and  {push task representations back to their initial task-optimal working region. Note that in this work we focus on adding more compact beneficial perturbations to the activation space, as adding perturbations to the input space has already been explored in adversarial attack methods, and adding perturbations to the parameter space is unlikely to be scalable due to the very large number of parameters in a typical neural network.}

\section{Beneficial Perturbation Network}

\label{bpn}


We implemented two variants of BPN: BD + EWC and BD + PSP (Experiments). The backbone - BD  (updating extra out-of-network bias units in beneficial directions to create beneficial perturbations) is the same for both methods. The only difference is BD + EWC (BD + PSP) uses EWC (PSP) method to retrain the normal weights while attempting to minimize disruption of old tasks. Here, we choose BD + EWC to explain our method (for BD + PSP, see Supplementary). We use a scenario with two tasks for illustration; task A - recognizing MNIST digit 1s, 2s, task B - recognizing MNIST digit 3s, 4s. BPN has task-dependent bias units ($\mathbf{BIAS}_{t}^{i}\in R^{1{\times}K}$, K is the number of normal neurons in each layer, $i$ is the layer number, and $t$ is the task number) in each layer to store the beneficial perturbations. The beneficial perturbations are formulated as an additive contribution to  each layer's weighted activations. Unlike most adversarial perturbations, beneficial perturbations are not specific to each example, but are applied to all examples in each task (Fig.~\ref{fig:beneficial_perturbations} c, d). We define beneficial perturbations as a task-dependent bias term:

\begin{equation} \mV^{i+1} = \sigma(\mW^{i}\mV^{i} +b^i  +\mathbf{BIAS}_{t}^{i} ) \ \ \ \mathbf{\forall} \ \ i \in [1,n] \label{Eqn:activations_rules} \end{equation} 

\noindent where $V^{i}$ is the activations at layer $i$, $W^{i}$ is the normal weights at layer $i$, $BIAS_{t}^{i}$ is the task dependent bias units at layer $i$ for task $t$, $\sigma(\cdot)$ is the nonlinear activation function at each layer, $b^i$ is the normal bias term at layer $i$, $n$ is the number of layers.
For a simple fully connected network (Fig.~\ref{fig:explanation_structure} a), the forward functions are:

\begin{equation}
    \mV^{1} = \sigma(\mW^{1}\mX_{t}+ b^1 +\mathbf{BIAS}_{t}^{1} )
\end{equation}
\begin{equation}
    \mV^{2} = \sigma(\mW^{2}\mV^1+b^2 +\mathbf{BIAS}_{t}^{2})
\end{equation}
\begin{equation}
       \mathbf{y} = Softmax(\mW^{3}\mV^2+ b^3 +\mathbf{BIAS}^{3}_{t})     
\end{equation}

\noindent where $\mathbf{y}$ is the output logits, $\mX_{t}$ is the input data for task t, $Softmax$ is the normalization function, other notations are the same as in Eqn.~\ref{Eqn:activations_rules}. During the training of a specific task, the bias units are the product of two terms\footnote{the factorization provides more degrees of freedom to better learn the beneficial perturbations \citep{haeffele2017global,du2019gradient}}: $\mM_{t}^{i}\in R^{1{\times}H}$ and $\mW_{t}^{i}\in R^{H{\times}K}$ (H is the hidden dimension (a hyper-parameter), K is the number of normal neurons in each layer, and $t$ is the task number). After training a specific task, we discard both $\mM_{t}^{i}$ and $\mW_{t}^{i}$, and only keep their product $\mathbf{BIAS}_{t}^{i}$, reducing memory and parameter costs to a negligible amount (0.3$\%$ increase for parameters per task, and 4*K Bytes increase per layer per task, it is just a bias term). After training on different sequential tasks, at test time, the stored beneficial perturbations from the specific bias units can bias the neural network outputs to each task. Thus, these allow the BPN to switch into different modes to process different tasks. We use the forward and backward rules (Alg.~\ref{alg:FORTA}, Alg.~\ref{alg:BACTA}) to update the BPN.


 {{\bf{For training}}}, first, during the training of task A, our goal is to maximize the probability $P(\mathbf{y} = \mathbf{y}_{A}|\mX_{A},\mW^{i},\mathbf{BIAS}_{A}^{i}) \ \ \ \mathbf{\forall} \ \ i \in [1,n]$ by selecting the bias units corresponding to tasks A . Thus, we set up our optimization function as: 

\begin{equation}
\small
\begin{aligned}
\mW^{i},&\,\mathbf{BIAS}_{A}^{i} = 
{ \mathop{\arg\min}_{ \mW^{i},\,\mathbf{BIAS}_{A}^{i}}} \\
 & { -\, log\,[\,P(\mathbf{y} = \mathbf{y}_{A}|\mX_{A},\,\mW^{i},\,\mathbf{BIAS}_{A}^{i})\,]}
\ \ \ \ \mathbf{\forall} \ \ i \in [1,n]
\end{aligned}
\label{Eqn:training_A} \end{equation}

\noindent where $\mathbf{y}_{A}$ is the true label for data in task A (MNIST input images 1, 2), $X_{A}$ is the data for task A, other notations are the same as notations in Eqn.~\ref{Eqn:activations_rules}. We update $\mM_{A}^{i}$ in the beneficial direction (FGSD) as $\epsilon sign(\nabla_{\mM_{A}^{i}} L(\mM_{A}^{i},\vy_{A}))$ to generate beneficial perturbations for task A, where $\mM_{A}^{i}$ are the first term of bias units for task A.  We update $\mW_{A}^{i}$ (the second term of bias units for task A) in the gradient direction. The factorization allows the bias units for task A to better learn the beneficial perturbations for task A (a vector towards the work space of task A that has non-negligible network response for MNIST digits 1, 2, similar to Fig.~\ref{fig:beneficial_perturbations}b, c ).  {We use a softmax cross entropy loss to optimize Eqn.~\ref{Eqn:activations_rules}.} After training task A, the bias units for task A ($\mathbf{BIAS}_{A}^{i}$) are the product of $\mM_{A}^{i}$ and $\mW_{A}^{i}$. We discard $\mM_{A}^{i}$ and $\mW_{A}^{i}$ to reduce the memory storage and parameter costs and freeze the $\mathbf{BIAS}_{A}^{i}$ to ensure that the beneficial perturbations are not being corrupted by other tasks (Task B). Then, we discard all of the MNIST input images 1, 2 because all of the information is stored inside the bias units for task A and we do not need to replay these images when we train on the following sequential tasks.  

After training task A, during the training of task B (Fig.~\ref{fig:explanation_structure} d), our goal is to maximize the 
probability ${P(\mathbf{y} = \mathbf{y}_{B}|\mX_{B},\mW^{i},\mathbf{BIAS}_{B}^{i})}$ ${  \mathbf{\forall}  \ i \in [1,n]}$ by selecting the bias units corresponding to tasks B. To minimize the disruption for task A, we apply EWC or PSP constraints on normal weights. We set up our optimization function as 

\begin{equation}
\small
\begin{aligned}
\mW^{i},&\,\mathbf{BIAS}_{B}^{i} =  \mathop{\arg\min}_{ \mW^{i},\,\mathbf{BIAS}_{B}^{i}} \\
 & -\, log\,[\,P(\mathbf{y} = \mathbf{y}_{B}|\mX_{B},\,\mW^{i},\,\mathbf{BIAS}_{B}^{i})\,]
 + EWC(\mW^{i}) \\
 & \ \ \ \ \ \ \ \ \ \ \ \ \ \ \ \ \ \mathbf{\forall} \ \ i \in [1,n]
\end{aligned}
\label{Eqn:training_EWC_constrained}\end{equation}
where $\mathbf{y}_{B}$ is the true label for data in task B (MNIST input images 3, 4), $X_{B}$ is the data for task B, $EWC(\cdot)$ is the EWC constraint \citep{kirkpatrick2017overcoming} on normal weights, other notations are the same as in Eqn.~\ref{Eqn:activations_rules}. In the loss function of Alg.~\ref{alg:BACTA}, $\lambda F_j(W_j-W^{A*}_{j})^2$ is the EWC constraint on the normal weights,  where $j$ labels each parameter, $F_j$ is the Fisher information matrix for each parameter $j$  {(determine which parameters are most important for a task \citep{kirkpatrick2017overcoming})}, $\lambda$ sets how important the old task is compared to the new one, $W_j$ is normal weight $j$, and $W_{j}^{A*}$ is the optimal normal weight $j$ after training on task A. Apart from the additional EWC constraint, training task B and all subsequent tasks then simply proceeds in the same manner as for task A above.

 {{{\bf For testing}}},  after training task B, we test the accuracy for task A on a test set by manually activating the bias units corresponding to task A (Fig.~\ref{fig:explanation_structure} c, Alg.~\ref{alg:FORTA}). Although the shared normal weights have been contaminated by task B, the integrity of bias units for task A that store the beneficial perturbations still can bias the network outputs to task A (set the network into a mode to process input from task A, see Results).  In another word, the task-dependent bias units can still maintain a high probability ~--~ $P(\mathbf{y} = \mathbf{y}_{A}|\mX_{A},\mW^{i},\mathbf{BIAS}_{A}^{i})$ for task A. During testing of task B, we test the accuracy for task B on a test set by manually activating the bias units corresponding to task B (Fig.~\ref{fig:explanation_structure} e, Alg.~\ref{alg:FORTA}). The bias units for task B can bias the network outputs to task B and maintain a high probability ~--~ $P(\mathbf{y} = \mathbf{y}_{B}|\mX_{B},\mW^{i},\mathbf{BIAS}_{B}^{i})$ for task B, in case the shared normal weights are further modified by later tasks. In scenarios with more than two tasks, the forward and backward algorithms for later tasks are the same as for task B, except that they will select and update their own bias units.

In sum, beneficial perturbations act upon the network not by adding biases to the input data (like adversarial examples do, Fig.~\ref{fig:beneficial_perturbations}a), but instead by dragging the drifted activations back to the correct working region in activation space for the current task ( Fig.~\ref{fig:concept_in_loss_space} and Fig. ~\ref{fig:beneficial_perturbations} c).
The intriguing properties of task-dependent beneficial perturbations on maintaining high probabilities for different tasks can further be explained in two ways. The beneficial perturbations from the bias units can be viewed as features that capture how "furry" the images are for task A (or B). Olshausen {\em et al.} \citep{cheung2019superposition} showed that training a neural network only on these features is sufficient to make correct classification on the dataset that generates these features. They argued that these features have sufficient information for a neural network to make correct classification. In our continual learning scenarios, although the shared normal weights ($\mW^{i}$) have been contaminated after the sequential training of all tasks, by activating corresponding bias units, the task-dependent bias units still have sufficient information to bias the network toward that task. In other words, the task-dependent bias units can maintain high probabilities ~--~ $P(\mathbf{y} = \mathbf{y}_{A}|\mX_{A},\mW^{i},\mathbf{BIAS}_{A}^{i})$ for task A or $P(\mathbf{y} = \mathbf{y}_{B}|\mX_{B},\mW^{i},\mathbf{BIAS}_{B}^{i})$ for task B . Thus, bias units can assist the network to make correct classification. In addition, Elsayed {\em et al.} \cite{elsayed2018adversarial} showed how a carefully computed  adversarial perturbations for each new task embedded in the input space can repurpose machine learning models to perform a new task. Here, these beneficial perturbations can be viewed as task-dependent beneficial "programs"\cite{elsayed2018adversarial} in the parameter space. Once activated, these task-dependent "programs" can maximize the probability for corresponding tasks.

\begin{algorithm*}[h]
\small
\caption{BD + EWC : forward rules for task t }
\label{alg:FORTA}

\begin{algorithmic}
\State {\quad For each fully connected layer i}
\State {\quad Select bias units (task t): $\mathbf{BIAS}_{t}^{i}$) for the current task }
\State{\quad{\bfseries Input:}\hspace{0.15cm} $\mathbf{BIAS}_{t}^{i}$ \textemdash
\hspace{0.08cm} Bias units for task t} \hfill {\color{green}// provide beneficial perturbations to bias the neural network}
\State{\quad\quad\quad\quad\hspace{0.33cm}$\mV^{i-1}$\textemdash
\hspace{0.08cm} Activations from the last layer}
\State{\quad{\bfseries Output:} $\mV^{i} = \sigma(\mW^{i} \cdot  \mV^{i-1} + \mathbf{b}^{i} +\mathbf{BIAS}_{t}^{i}) \ \ \ \mathbf{\forall}  \ i \in [1,n]$ \hfill {\color{green}// activations for the next layer}}
\State{\quad\quad\quad\qquad \hspace{0.13cm} where:  $\mW^{i}$\textemdash
\hspace{0.08cm} normal neuron weights at layer $i$. $\mathbf{b}^{i}$\textemdash \hspace{0.08cm} normal bias term at layer $i$}
\State{\quad\quad\quad\qquad \hspace{1.23cm} $n$ \textemdash \hspace{0.08cm} the number of FC layers. \hspace{1.1cm} $\sigma(\cdot)$ \textemdash \hspace{0.08cm} the nonlinear activation function at each layer}
\end{algorithmic}
\end{algorithm*}

\begin{algorithm*}[h]
\small
\caption{BD + EWC : backward rules for task t }
\label{alg:BACTA}

\begin{algorithmic}

\State {{\underline {For the first task A ($t = 1$):}}}
\State{{\quad Minimizing loss function: $L(\mX_{A},\mW^{i},\mathbf{BIAS}_{A}^{i})  \ \ \ \mathbf{\forall}  \ i \in [1,n]$}}
\State{{\quad \quad where: $\mX_{A}$\textemdash
\hspace{0.08cm} data for task One.\quad $\mW^{i}$\textemdash
\hspace{0.08cm} normal neuron weights at layer $i$. }}
\State{{  \quad\quad\quad\quad \hspace{0.16cm} $\mathbf{BIAS}_{A}^{i}$ \textemdash
\hspace{0.08cm}  bias units for task One from FC layers $i$, which is the product of $(\mM_{A}^{i}, \mW_{A}^{i})$}} 
\State{{\hspace{0.16cm}\quad\quad\quad\quad\quad $n$\textemdash
\hspace{0.08cm} the number of FC layers.}}

\State{}

\State {\underline {For task B ($t > 1$):}}
\State{\quad Minimizing loss function: $L(\mX_{t},\mW^{i}, \mathbf{BIAS}_{B}^{i})+ \sum_{j} \lambda F_j(W_j-W^{A*}_{j})^2  \ \ \ \mathbf{\forall}  \ i \in [1,n]$}
\State{{\quad \quad where: $\mX_{B}$\textemdash
\hspace{0.08cm} data for task B.\quad $\mW^{i}$\textemdash
\hspace{0.08cm} normal neuron weights from FC at layers $i$}}
\State{{\hspace{0.1cm}\quad\quad\quad\quad\quad $j$\textemdash
\hspace{0.08cm} labels each parameter.\quad $F_{j}$  \textemdash
\hspace{0.08cm} Fisher information matrix for parameter j.}}
\State{{\hspace{0.1cm}\quad\quad\quad\quad\quad $W_j$\textemdash
\hspace{0.08cm} normal weight j.\hfill $W^{A*}_{j}$  \textemdash
\hspace{0.08cm} optimal normal weight j after training on task A.}}
\State{{  \quad\quad\quad\quad \hspace{0.16cm} $\mathbf{BIAS}_{B}^{i}$ \textemdash
\hspace{0.08cm}  bias units for task B at FC layers $i$, which is the product of $(\mM_{B}^{i}, \mW_{B}^{i})$}} 
\State{{\hspace{0.1cm}\quad\quad\quad\quad\quad $n$\textemdash
\hspace{0.08cm} the number of FC layers.}}
\State{}
\State{\bf \underline{For each fully connected layer i:}}
\State{}
\State{\quad \underline {During the training of task t}}
\State {\quad \hspace{0.07cm} Select bias units for the current task t ($\mathbf{BIAS}_{t}^{i}$)}
\State{\hspace{0.07cm} \quad{\bfseries Input:}\hspace{0.15cm} $\mathbf{Grad}$ \textemdash
\hspace{0.08cm} Gradients from the next layer}
\State{\hspace{0.07cm} \quad{\bfseries output:} $\mathbf{dW_{t}^{i}} = \mathbf{Grad}\cdot((\mM_{t}^{i})^T)$  {\color{green}// gradients for the second term of bias units for task t at layer i } }
\State{\hspace{0.05cm}  \quad  $\hspace{33pt}$   $\mathbf{dM_{t}^{i}} = \epsilon\; sign\;((\mathbf{W}_{t}^{{i}})^T\cdot(\mathbf{Grad}))$ }
\State{\hspace{0.07cm} \hfill  {\color{green}// gradients for the first term of bias units for task t at layer i using FGSD method}}
\State{\hspace{0.08cm} \quad   $\hspace{33pt}$   $\mathbf{dW^i} = \mathbf{Grad}\cdot((\mathbf{V}^{i})^T)$\hfill {\color{green}// gradients for normal weights at layer i}}
\State{\hspace{0.08cm} \quad  $\hspace{33pt}$   $\mathbf{dV^{i}} = (\mathbf{W}^{i})^T \cdot (\mathbf{Grad})$ \hfill {\color{green}// gradients for activations at layer i to last layer i -1}}
\State{\hspace{0.07cm} \quad  $\hspace{33pt}$   $\mathbf{db^i} =  \sum_{j} {Grad}_j $ \hfill {\color{green}// gradients for normal bias at layer i, j is iterator over the first dimension of {\bf{Grad}} }}
\State{}
\State {\quad \underline {After training of task t}}
\State{\quad \hspace{0.07cm}  Freeze the $\mathbf{BIAS}_{t}^{i}$}
\State{\quad \hspace{0.07cm}  Delete the $\mathbf{W_{t}^{i}}$ and  $\mathbf{M_{t}^{i}}$ to reduce parameter and memory storage cost}

\end{algorithmic}
\end{algorithm*}

\section{Experiments}
\label{experiments}
\subsection{Experimental Setup For Incremental Tasks} 
To demonstrate that BPN is very parameter efficient and  can learn different tasks in an online and continual manner, we used a fully-connected neural network with 5 hidden layers of 300 ReLU units. We tested it on three public computer vision datasets with "single-head evaluation", where the output space consists of all the  classes from all tasks learned so far.  

{\bf 1. Incremental MNIST.} A variant of the MNIST dataset \citep{lecun1998gradient} of handwritten digits with 10 classes, where each task introduces a new set of classes. We consider 5 tasks; each new task concerns examples from a disjoint subset of 2 classes. 

{\bf 2. Incremental CIFAR-10.} A variant of the CIFAR object recognition dataset \citep{krizhevsky2009learning} with 10 classes. We consider 5 tasks; each new task has 2 classes. 

{\bf 3. Incremental CIFAR-100.} A variant of the CIFAR object recognition dataset \citep{krizhevsky2009learning} with 100 classes. We consider 10 tasks; each new task has 2 classes. We use 20 classes for CIFAR-100 experiment.

\subsection{Experimental Setup For Eight Sequential Object Recognition Tasks}

To demonstrate the superior performance of BPN across different datasets and domains, we consider a sequence of eight object recognition datasets: {\bf 1.} Oxford \textit{Flowers} \citep{nilsback2008automated} for fine-grained flower classification (8,189 images in 102 categories); {\bf 2.} MIT \textit{Scenes}  \citep{quattoni2009recognizing} for indoor scene classification (15,620 images in 67 categories); {\bf 3.} Caltech-UCSD \textit{Birds} \citep{wah2011caltech} for fine-grained bird classification (11,788 images in 200 categories); {\bf 4.} Stanford \textit{Cars} \citep{krause20133d} for fine-grained car classification (16,185 images of 196 categories); {\bf 5.} FGVC-\textit{Aircraft} \citep{maji2013fine} for fined-grained aircraft classification (10,200 images in 70 categories); {\bf 6.} VOC \textit{actions} \citep{everingham2015pascal}, the human action classification subset of the VOC challenge 2012 (3,334 images in 10 categories); {\bf 7.} \textit{Letters}, the Chars74K datasets \citep{de2009character} for character recognition in natural images (62,992 images in 62 categories); and {\bf 8.} the Google Street View House Number \textit{SVHN} dataset \citep{netzer2011reading} for digit recognition (99,289 images in 10 categories). To have a fair comparison,  we use the same  AlexNet \citep{krizhevsky2012imagenet} architecture pretrained on ImageNet \citep{russakovsky2015imagenet} as Aljundi {\em et al.} \cite{aljundi2018selfless,aljundi2018memory}, and tested on 8 sequential tasks with a "multi-head evaluation", where each task has its own classification layer (introduce same parameter costs for every method) and output space. All methods have a task oracle at test time to decide which classification layer to use. We run the different methods on the following sequence: Flower -> Scenes -> Birds -> Cars -> Aircraft -> Action -> Letters -> SVHM. 

\subsection{Experimental Setup for 100 permuted MNIST dataset}
To demonstrate that BPN has capacity to accommodate a large number of tasks, we tested it on 100 permuted MNIST datasets generated from randomly permuted handwritten MNIST digits. We consider 100 tasks; each new task has 10 classes. We used a fully-connected neural network with 4 hidden layers of 128 ReLu Units (a core network with small capacity) to compare the performances of different methods. The type 4 methods, such as the Parameter Superposition (PSP \citep{cheung2019superposition}) would exhaust the unrealized capacity and inevitably dilute the capacity of the core network under a large number of tasks: in their Fig. 2, with a network that has 128 hidden units (leftmost panel), the average task performance for all tasks trained so far, is 95\% after training one task, but decreases to 50\% after training fifty tasks. While a larger network with 2048 hidden units shows much smaller decrease from 96\% to about 90\% (see Fig. 2 in their paper, rightmost panel). The reason is that this method generates a random diagonal binary matrix for each task, which in essence is a key or selector for that task. As more and more tasks are learned, those keys start to overlap more, causing interference among tasks. In comparison, BPN can counteract the dilution, hence it can accommodate a large number of tasks.

\subsection{Our model and baselines}  We compared the proposed Beneficial Perturbation Network ( Beneficial Perturbation + Elastic Weight Consolidation (the eleventh model), BD + EWC (variant 1) and Beneficial Perturbation + Parameter Superposition (the twelfth model), BD + PSP (variant 2)) to 11 alternatives to demonstrate its superior performance.

{\bf 1. Single Task Learning (STL).} We consider several 5-layer fully-connected neural networks. Each network is trained for each task separately. Thus, STL does not suffer from catastrophic forgetting at all. It is used as an upper bound.

{\bf 2. Elastic Weight Consolidation (EWC) \citep{kirkpatrick2017overcoming}.} The loss is regularized to avoid catastrophic forgetting.

{\bf 3. Gradient Episodic Memory with task oracle (GEM (*)) \citep{lopez2017gradient},} GEM uses a task oracle to build a final linear classifier (FLC) per task. The final linear classifier adapts the output distributions to the subset of classes for each task. GEM uses an episodic memory to store a subset of the observed examples from previous tasks, which are interleaved with new data from the latest task to produce a new classifier for all tasks so far. We use notation GEM (*) for the rest of the paper, where * is the size of episodic memory (number of training images stored) for each class.

{\bf 4. Incremental Moment Matching \citep{lee2017overcoming} (IMM)} IMM incrementally matches the moment of the posterior distribution of the neural network with a L2 penalty and equally applies it to changes to the shared parameters.

{\bf 5. Learning without forgetting \citep{li2017learning} (LwF)} First, LwF freezes the shared parameters while learning a new task. Then, LwF trains all the parameters until convergence.

{\bf 6. Encoder based lifelong learning \citep{rannen2017encoder} (EBLL)} Based on LwF, using an autoencoder to capture the features that are crucial for each task.

{\bf 7. Synaptic Intelligence \citep{zenke2017continual} (SI)} While training on new task, SI estimates the importance weights in an online manner. Parameters important for previous tasks are penalized during the training of new task.

{\bf 8. Memory Aware Synapses \citep{aljundi2018memory} (MAS)} Similar to SI method, MAS estimates the importance weights through the sensitivity of the learned function on training data. Parameters important for previous tasks are penalized during the training of new task.

{\bf 9. Sparse coding through Local Neural Inhibition and Discounting \citep{aljundi2018selfless} (SLNID)} SLNID proposed a new regularizer that penalizes neurons that are active at the same time to create sparse and decorrelated representations for different tasks. 

{\bf 10. Parameter Superposition  \citep{cheung2019superposition} (PSP)} PSP used task-specific context matrices to map different inputs from different tasks to different subspaces spanned by rows in the weight matrices to avoid interference between tasks. We use the binary superposition model of PSP throughout the paper, because it is not only more memory efficient, but also, in our testing, it performed better than other PSP variants (e.g., complex superposition).

{\bf 11. BD + EWC (ours):} Beneficial Perturbation Network (variant 1). The first term ($\mM_{t}$) of the bias units is updated in the beneficial direction (BD) using FGSD method. The second term ($\mW_{t}$) of the bias units is updated in the gradient direction. The normal weights are updated with EWC constraints.  

{\bf 12. BD + PSP (ours):} Beneficial Perturbation Network (variant 2). The first term ($\mM_{t}$) of the bias units is updated in the beneficial direction (BD) using FGSD method. The second term ($\mW_{t}$) of the bias units is updated in the gradient direction. The normal weights are updated using PSP (binary superposition model, Supplementary).  

\begin{figure}[]
	\begin{center}
		\includegraphics[width=0.85\linewidth]{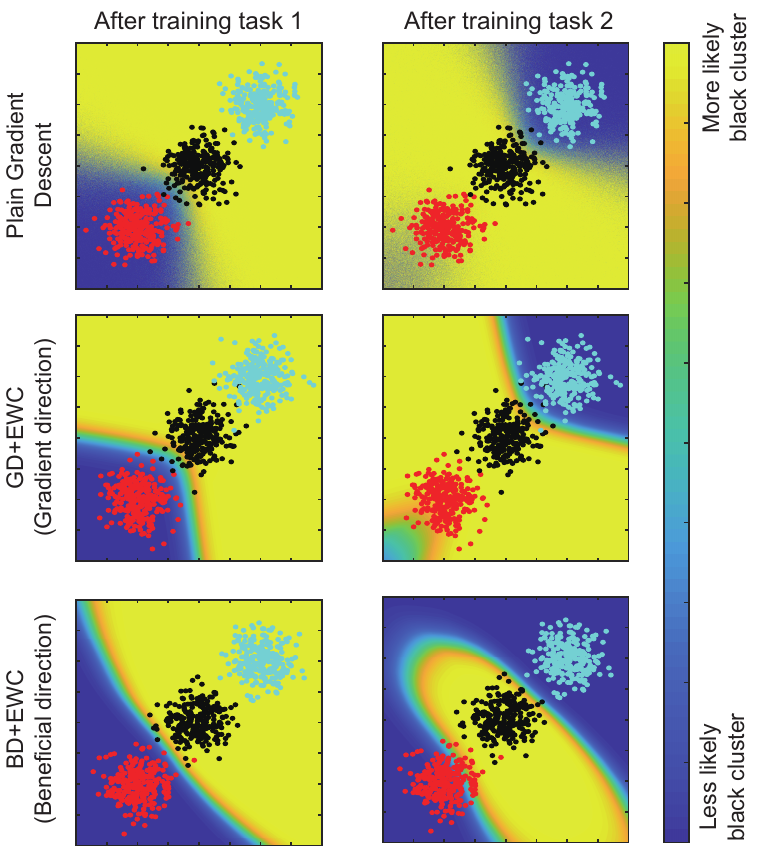}
		\caption{{\bf Visualization of classification regions:} classify 3 randomly generated normal distributed clusters. Task 1: separate black from red clusters. Task 2: separate black from light blue clusters. The yellower (bluer) the heatmap, the higher (lower) the chance the neural network classifies a location as the black cluster. After training task 2, only BD + EWC remembers task 1 by maintaining its decision boundary between the black and red clusters. Both plain gradient descent and GD + EWC forget task 1 entirely.}
		\label{fig:visualization}
	\end{center}
\end{figure}

{\bf 13. GD + EWC:} The update rules and network structure are the same as BD + EWC, except the first term ($\mM_{t}$) of the bias units is updated in the Gradient direction (GD). This method has the same parameter costs as BD + EWC . The failure of GD + EWC suggests that the good performance of BD + EWC is not from the additional dimensions provided by bias units.

\section{ Results:}
\label{results}
\subsection{The beneficial perturbations can bias the network and maintain the decision boundary}  To show the advantages of our method are really from the beneficial perturbations and not just from additional dimensions to the neural network, we compare between updating the first term of the bias units in the beneficial direction (BD + EWC which comes from beneficial perturbations) and in the gradient direction (GD + EWC, which just comes from the additional dimensions that our bias units provide). We use a toy example (classifying 3 groups of Normal distributed clusters) to demonstrate it and to visualize the decision boundary (Fig.~\ref{fig:visualization}). We randomly generate 3 normal distributed clusters different locations. We have two tasks - Task 1: separate the black cluster from the red cluster. Task 2: separate the black cluster from the light blue cluster. The yellower (bluer) the heatmap, the higher (lower) the confidence that the neural network classifies a location into the black cluster. After training task 2, both plain gradient descent and GD + EWC forget task 1 (dark blue boundary around the red cluster disappeared). However, BD + EWC not only learns how to classify task 2 (clear decision boundary between light blue and black clusters), but also remembers how to classify the old task 1 (clear decision boundary between red and black clusters). Thus, the beneficial perturbations are what can bias the network outputs and maintain the decision boundary for each task, not just adding more dimensions.

\begin{figure*}[h]
	\begin{center}
		\includegraphics[height=10.8cm]{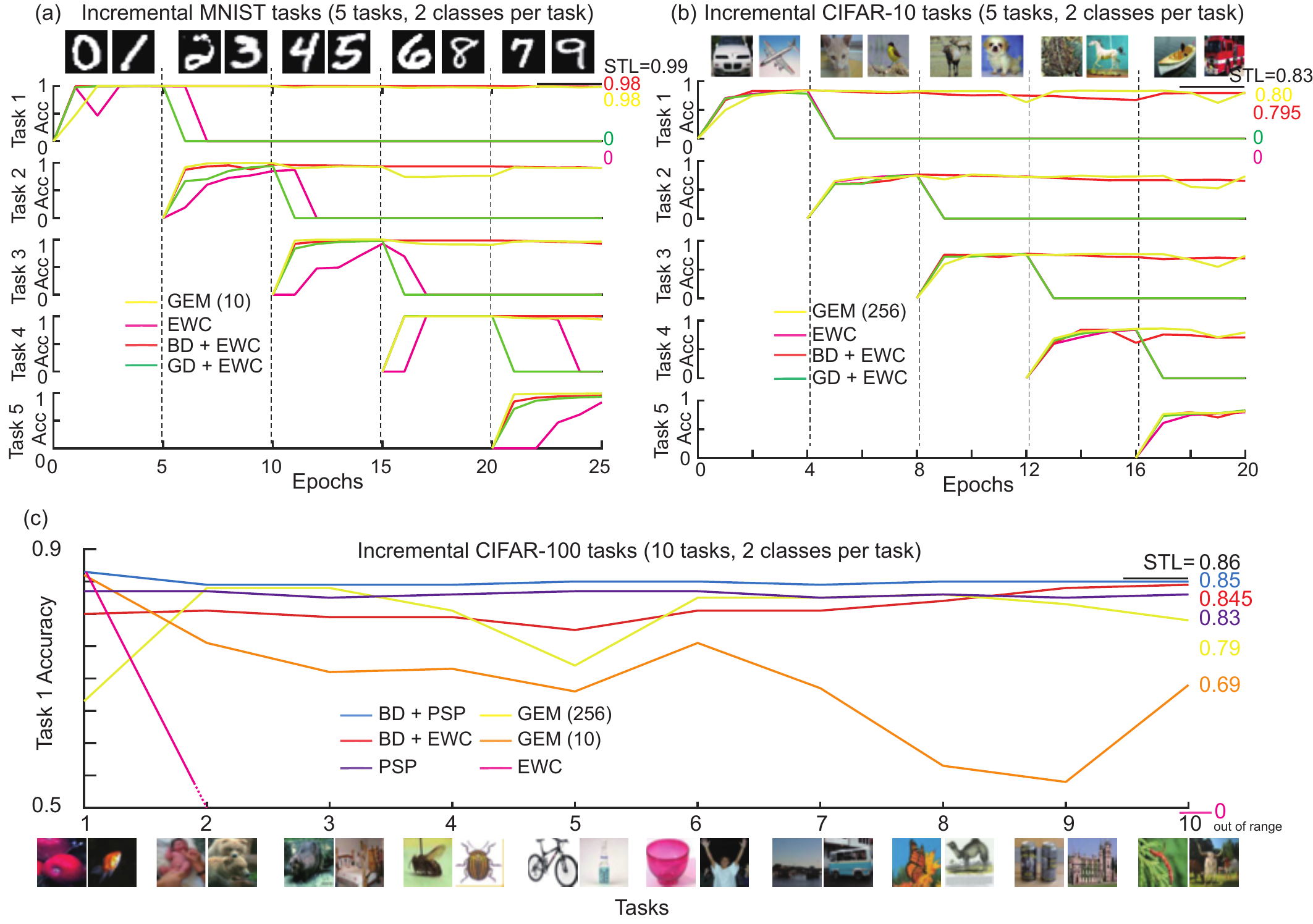}
		\caption{ Results for a fully-connected network with 5 hidden layers of 300 ReLU units. (a) Incremental MNIST tasks (5 tasks, 2 classes per task). (b) Incremental CIFAR-10 tasks (5 tasks, 2 classes per task). For a and b, the dashed line indicates the start of a new task. The vertical axis is the accuracy for each task. The horizontal axis is the number of epochs. (c) Incremental CIFAR-100 tasks (10 tasks, 2 classes per task). The vertical axis is the accuracy for task 1. The horizontal axis is the number of tasks.}
		\label{fig:quantative_results}
	\end{center}
\end{figure*}

\begin{table*}[h!]
\caption{Task 1 performance with "single-head" evaluation after training all sequential tasks on incremental MNIST, CIFAR-10 and CIFAR-100 Dataset. We include additional memory storage costs per task  (extra components that are necessary to be stored onto the disks after training each task, Supplementary) of GEM , BD+EWC, BD + PSP and PSP method.}
\label{tab:memory_performance}
\vskip 0.15in
\begin{center}
\begin{small}
\begin{sc}
\begin{tabular}{cccl}
\toprule
Dataset & Method & \makecell{Task 1 performance after\\ training all sequential tasks} & \makecell{additional memory storage  \\ costs per task (Bytes)}\\
\midrule
\makecell{Incremental MNIST\\ (5 tasks, 2 classes per task)} & \makecell{GEM(10)\\BD+EWC}& \makecell{0.980 \\\bf{0.980}}& \makecell[r]{\ \ \ \ \ 47,040 \\\bf{4,808} }\\\hline

\makecell{Incremental CIFAR-10\\ (5 tasks, 2 classes per task)} & \makecell{GEM(256)\\GEM(150)\\BD+EWC}& \makecell{\bf{0.800} \\0.698\\0.795}& \makecell[r]{4,718,592  \\2,764,800 \\\bf{4,808} }\\\hline

\makecell{Incremental CIFAR-100\\ (10 tasks, 2 classes per task)} & \makecell{GEM(256)\\GEM(209)\\BD+PSP\\PSP\\BD+EWC}& \makecell{0.790\\0.775 \\ \bf{0.850}\\0.830\\0.845}& \makecell[r]{4,718,592 \\3,852,288 \\20,776\\15,968\\\bf{4,808} }\\ \hline

\end{tabular}

\end{sc}
\end{small}
\end{center}
\vskip -0.1in
\end{table*}

\subsection{Quantitative analysis for incremental tasks} Our BPN achieves a comparable or better performance than PSP, GEM, EWC, GD + EWC in "single-head" evaluations, where the output space consists of all the  classes from all tasks learned so far. In addition, it introduces negligible parameter and  memory storage costs per task. Fig.~\ref{fig:quantative_results} and Tab.~\ref{tab:memory_performance} summarize  performance for all datasets and methods. STL has the best performance since it trained for each task separately and did not suffer from catastrophic forgetting at all. Thus, STL is the upper bound. BD + EWC performed slightly worse than STL (1\%,4\%,1\% worse for incremental MNIST, CIFAR-10, CIFAR-100 datasets). BD + EWC achieved comparable or better performance than GEM. On incremental CIFAR-100 (10 tasks, 2 classes per task), BD + EWC outperformed PSP, GEM (256) and GEM (10) by 1.80\%, 6.96\%, and 22.4\%.  BD + PSP outperformed PSP, GEM (256) and GEM (10) by 2.40\%, 7.59\%, and 23.1\%.  By comparing the memory storage costs (Tab.~\ref{tab:memory_performance}, Supplementary), to achieve similar performance,  BD + EWC only introduces an additional 4,808 Bytes memory per task, which is only 0.1\% of the memory storage cost required by GEM (256). BD + PSP only introduces 20,776 Bytes, or 0.44\% of the memory storage cost required by GEM (256). 
The memory storage costs of BD + EWC is 30\% of that of PSP. The memory storage costs of BD + PSP is of the same order of magnitude as PSP. EWC alone rapidly decreased to 0\% accuracy. This confirms similar results on EWC performance on incremental datasets \citep{rios2018closed, kemker2017fearnet, parisi2019continual, kemker2018measuring} in "single-head" evaluations although EWC generally performs well in "multi-head" tasks. GD + EWC has the same additional dimensions as BD + EWC, but GD + EWC failed in the continual learning scenario. This result suggests that it is not the additional dimensions of the bias units, but the beneficial perturbations, which help overcome catastrophic forgetting.

\begin{table*}[h]
\caption{Test accuracy (in percent correct) achieved by each method with "multi-head" evaluation for each dataset after training on the 8 sequential object recognition datasets. (Dash (--) means that the results are not available in their papers. Star (*) means that we didn't reproduce the methods and the results were taken from SLNID \citep{aljundi2018selfless} and MAS \citep{aljundi2018memory}. Thus, we keep the same percentage table format as theirs).}
\label{tab:eighttasks}
\begin{center}
\includegraphics[width=0.97\linewidth]{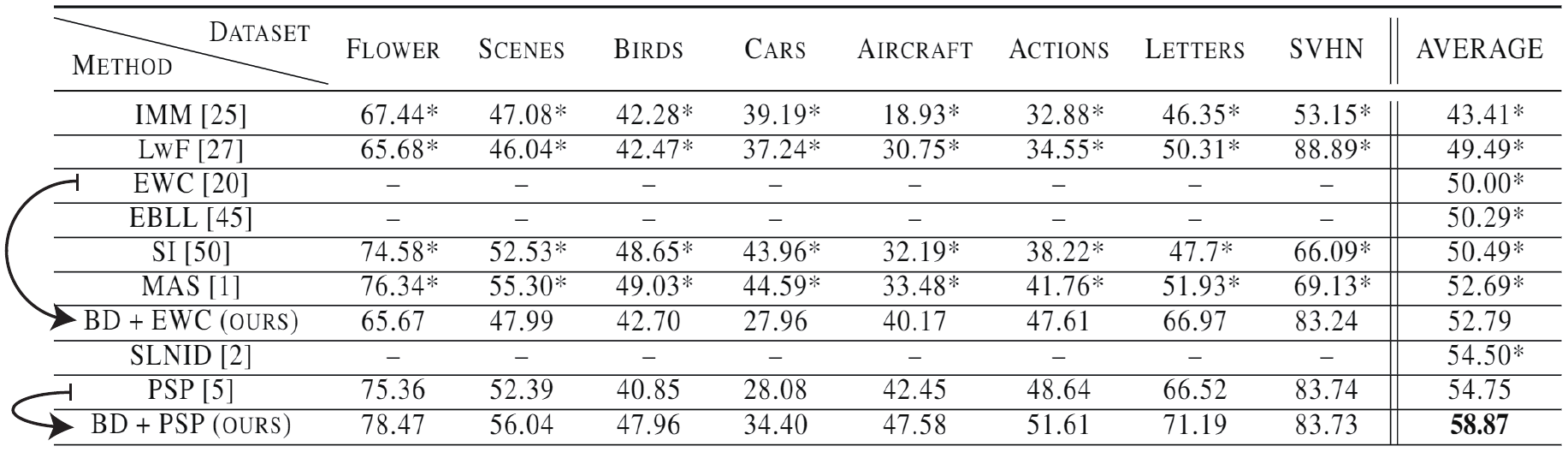}
\end{center}
\end{table*}

\subsection{Quantitative analysis for eight sequential object recognition tasks} The eight sequential object recognition tasks demonstrate the superior performance of BPN (BD + PSP or BD + EWC) compared to the state-of-the-art and the ability to learn sequential tasks across different datasets and different domains. Our BPN achieves much better performance than IMM \citep{lee2017overcoming}, LwF\citep{li2017learning}, EWC \citep{kirkpatrick2017overcoming}, EBLL \citep{rannen2017encoder}, SI \citep{zenke2017continual}, MAS \citep{aljundi2018memory}, SLNID \citep{aljundi2018selfless}, PSP \citep{cheung2019superposition} in "multi-head" evaluations, where each task has its own classification layer and output space. After training on the 8 sequential object recognition datasets, we measured the test accuracy for each dataset and calculated their average performance (Tab.~\ref{tab:eighttasks}). On average, BD + PSP (ours) outperforms all other methods: PSP (7.52\% better), SLNID (8.02\% better), MAS (11.73\% better), SI (16.60\% better), EBLL (17.07\% better), EWC (17.75\% better), LwF (18.96\% better) and IMM (35.62\% better). Although MAS, SI and EBLL performed better than EWC alone, with the help of our beneficial perturbations (BD), BD + EWC can achieve a better performance than these methods: MAS (0.34\% better), SI (4.71\% better), EBLL (5.13\% better) and EWC (5.74\% better). By including the BD (BD + PSP  and BD + EWC), we can significantly boost performance when compared to using PSP or EWC alone (black arrows in the Tab.~\ref{tab:eighttasks}).

\begin{figure}[h]
  \includegraphics[width=\columnwidth]{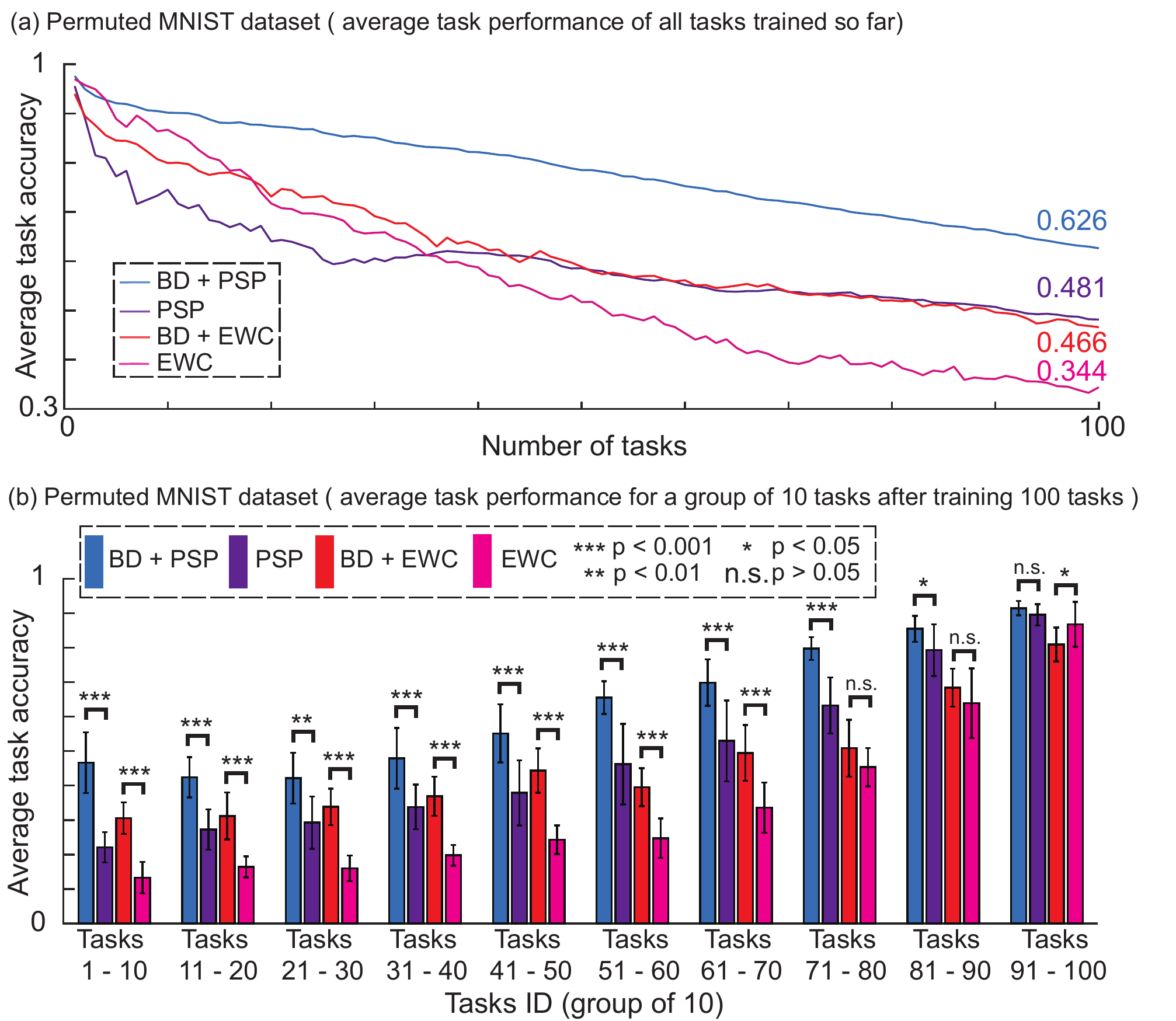}
  \caption{100 permuted MNIST datasets results for a fully-connected network with 4 hidden layers of 128 ReLU units. This network is relatively small for these tasks and hence does not offer much available redundancy or unrealized capacity. (a) The average task accuracy of all tasks trained so far as the number of tasks increases. (b) After training 100 tasks, the average task accuracy for a group 10 tasks. We use t-test to validate the results. \label{fig:100_permuted_MNIST_tasks}}
\end{figure}

\subsection{Quantitative analysis for 100 permuted MNIST dataset} 100 permuted MNIST dataset demonstrates that our BPN has capacity to accommodate a large number of tasks. After training 100 permuted MNIST tasks, the average task performance of BD + PSP is 30.14\% better than PSP. The average task performance of BD + EWC is 35.47\% higher than EWC (Fig.~\ref{fig:100_permuted_MNIST_tasks}.a). As the number of tasks increases (Fig.~\ref{fig:100_permuted_MNIST_tasks}.a), the average task performance of BD + PSP becomes increasingly better than PSP. The reason is that adding new tasks significantly dilutes the capacity of the original network in Type 4 methods (e.g., PSP) as there are limited routes or subspaces to form sub-networks. In this case, even though the core network can no longer fully separate each task, the Beneficial perturbations (BD) can drag the misrepresented activations back to their correct work space of each task and recover their separation (as demonstrated in Fig.~\ref{fig:beneficial_perturbations}). Thus, the BD of BD + PSP can still increase the capacity of the network and boost the performance. Similarly, BD components in BD + EWC can boost performance, increasing the capacity of the network to accommodate more tasks than EWC alone (Fig.~\ref{fig:100_permuted_MNIST_tasks}.b). In addition, after training 100 tasks (Fig.~\ref{fig:100_permuted_MNIST_tasks}. b), the accuracy of BD + EWC for the first 50 tasks is higher than PSP, likely because BD+EWC did not severely dilute the core network's capacity while PSP did. This means BD + EWC has a larger capacity than PSP. In contrast, the lower performance of the last 50 tasks for BD + EWC comes from the constraints of EWC (do not allow the parameters of the network learned from the new tasks to have large deviations from the parameters trained from old tasks). Although the performance of PSP is much better than EWC, with the help of BD, BD + EWC still reaches a similar performance as PSP.  

\section{Discussion}

\label{discussion}

We proposed a fundamentally new biologically plausible type of method - beneficial perturbation network (BPN), a neural network that can switch into different modes to process independent tasks, allowing the network to create potentially unlimited mappings between inputs and outputs. We successfully demonstrated this in the continual learning scenario.  Our experiments demonstrate the performance of BPN is better than the state-of-the-art. 1) BPN is more parameter efficient (0.3\% increase per task) than the various network expansion and network partition methods. it does not need a large episodic memory to store any data from previous tasks, compared to episodic memory methods, or large context matrices, compared to partition methods. 2) BPN achieves state-of-the-art performance across different datasets and domains. 3) BPN has a larger capacity to accommodate a higher number of tasks than the partition networks.  Through visualization of classification regions and quantitative results, we validate that beneficial perturbations can bias the network towards a task, allowing the network to switch into different modes. Thus, BPN significantly contributes to alleviating catastrophic forgetting  and achieves much better performance than other types of methods.

Elsayed {\em et al.} \cite{elsayed2018adversarial} showed how carefully computed  adversarial perturbations embedded in the input space can repurpose machine learning models to perform a new task without changing the parameters of the models. This attack finds a single adversarial perturbation for each task,  to cause the model to perform a task chosen by the adversary.
This adversarial perturbation can thus be considered as a program to execute each task. Here, we leverage similar ideas. But, in sharp contrast, instead of using malicious programs embedded in the input space to attack a system, we embedded beneficial perturbations ('beneficial programs') into the network's parameter space (the bias terms), enabling the network to switch into different modes to process different tasks. The goal of both approaches is similar - maximizing the probability ($P(current \ task|image\  input,\  program)$) of the current task given the image input and the corresponding program for the current task. This can be achieved by either forcing the network to perform an attack task in Elsayed {\em et al.}, or assisting it to perform a beneficial task in our method. The addition of programs to either input space (Elsayed {\em et al.}'s method) or the network's activation space (our method) helps the network maximize this probability for a specific task.  

We suggest that the intriguing property of the beneficial perturbations that can bias the network toward a task might come from the property of adversarial subspaces. Following the adversarial direction, such as by using the fast gradient sign method (FGSD) \cite{goodfellow6572explaining}, can help in generating adversarial examples that span a continuous subspace of large dimensionality (adversarial subspace). Because of “excessive linearity” in many neural networks \cite{tramer2017space} \cite{goodfellow2016}, due to features including Rectified linear units and Maxout, the adversarial subspace often takes a large portion of the total input space. Once an adversarial input lies in the adversarial subspace, nearby inputs also tend to lie in it. Interestingly, this corroborates recent findings by Ilyas {\em et al.} \citep{ilyas2019adversarial} that imperceptible adversarial noise can not only be used for adversarial attacks on an already-trained network, but also as features during training. For instance, after training a network on dog images perturbed with adversarial perturbation calculated from cat images, the network can achieve a good classification accuracy on the test set of cat images. This result shows that those features (adversarial perturbations) calculated from the cat training sets, contain sufficient information for a machine learning system to make correct classification on the test set of cat images. In our method, we calculate those features for each task, and store them into the bias units. In this case, although the normal weights have been modified (information from old tasks are corrupted), the stored beneficial features for each task have sufficient information to bias the network and enable the network to make correct predictions.

 {BPN is loosely inspired by its counterpart in the human brain: having task-dependent modules such as bias units in our Beneficial Perturbation Network, and long-term memories in hippocampus (HPC, \cite{bakker2008pattern}) in a brain network, are crucial for a system to switch into different modes to process different tasks. During weight consolidation, the HPC \citep{lesburgueres2011early,squire1995retrograde,frankland2005organization,helfrich2019bidirectional} fuses features from different tasks into coherent memory traces. Over days to weeks, as memories mature, the HPC progressively stores permanent abstract high-level long-term memories to remote memory storage areas (neocortical regions). The HPC can then maintain and mediate their retrieval independently when a specific memory is in need. 
We suggest that when a specific memory is retrieved, it helps the HPC switch into distinct modes to process different tasks. 
Thus, our analogy between HPC and BPN can be formulated as: during the training of BPN, updating the shared normal weights using EWC or PSP in theory leads to distinct task-dependent representations (similar to the coherent memory traces in HPC). However, some overlap between these representations is inevitable because model parameters become too constrained for EWC, or PSP  runs out of unrealized capacity of the core network. To circumvent this effect, Bias units (akin to the long-term memories in the neocortical areas) are trained independently for each task. At test time, bias units for a given task are activated to push representations of old tasks back to their initial task-optimal working regions in an analogous manner to maintaining and mediating the retrieval of Long-term memories independently in HPC.}

 {An alternative biological explanation evokes the concept of factorized codes. In biological neuronal populations,  neurons can be active for one task or, in many cases, for more than one tasks. At the population level, different tasks are encoded by different neuronal ensembles which can overlap. In our model, the PSP component deploys binary keys to activate task-specific readouts in hidden layers, in an analogy to neuronal task ensembles. When activating a BD component for a task, we would be further disambiguating a task-specific ensemble, particularly across neurons which are active for more than one task.  The reason for this is that adding task-specific beneficial perturbations to activations of hidden layers can shift the distribution of the net activation (akin to a DC offset or carrier frequency). Evidence from nonhuman primate experiments \citep{roy2010prefrontal,cromer2010representation} and human behavioral results \citep{flesch2018comparing} support this factorized code theory.  Electrophysiological experiments using monkeys demonstrated that neurons in prefrontal cortex are either representing competing categories independently \citep{roy2010prefrontal} or could represent multiple categories \citep{cromer2010representation}. In human behavior experiments, "humans tend to form factorized representation that optimally segregated the tasks \citep{flesch2018comparing}". In addition, recent neural network simulations \citep{yang2019task} demonstrated that "network developed mixed task selectivity similar to recorded prefrontal neurons after learning multiple tasks sequentially with a continual learning technique". Thus, having factorized representations for different tasks is important for enabling life-long learning and designing a general adaptive artificial intelligence system. }

\section*{Acknowledgment}
This work was supported by the National Science Foundation (grant number CCF-1317433), C-BRIC (one of six centers in JUMP, a Semiconductor Research Corporation (SRC) program sponsored by DARPA), and the Intel Corporation. The authors affirm that the views expressed herein are solely their own, and do not represent the views of the United States government or any agency thereof.



%
\bibliographystyle{abbrv}
\bibliography{bare_conf}

\section{Supplementary}

\subsection{Clarification of memory storage costs}

For the memory storage costs, we only consider what components are necessary to be stored on the disk after training each task, that is, the cost of storing the model for later re-use. In other words, the memory storage cost is defined as “the number of bytes required to store all of the parameters in the trained model" \citep{iandola2016squeezenet} (in table 2 of Iandola et al.).  This is usually the metric that is reported along with the number of operations to run a model (e.g., mobilenet web page, darknet web page, SqueezeNet \citep{iandola2016squeezenet} and Additive Parameter Superposition \citep{yoon2019oracle}). The extra memory storage costs of EWC are zero under this definition. Indeed, even though it requires a lot of transient memory during training, in the end the contents of this memory are used only to constrain network weight updates, and they are discarded once a training run is complete. At test time, a network trained with EWC has the same number of parameters, uses the same amount of runtime memory, and the same amount of operations as the original model. For example, consider that after training 5 sequential tasks, we want to train a new task 6. There are 5 steps: 1) Load the trained model from disk; 2) EWC would make a duplication of the parameters learned so far and just loaded from disk, and put them into transient memory (RAM); 3) During the training of task 6, EWC calculates the Fisher information matrix and applies the EWC constraints, which relies on the contents of the transient memory; 4) Delete the duplications of the parameters in the transient memory; 5) Save the parameters of latest model onto disk. In contrast, after training each task, GEM needs to store some images from that task onto the disk. Likewise, PSP needs to store the context matrix for each new task to disk, and BD + EWC needs to store the bias units to disk. Thus, the extra memory storage costs of BD + EWC is just the memory storage costs of the bias units (BD).

 {\subsection{Clarification of parameter costs}}

 {Similar to memory storage costs, we only consider what components are necessary to be stored on disk after training each task (0.3\% increase per task), that is, the cost of storing the model for later re-use. It should be noted that BPN needs large additional weight matrices (called $W^i_t$ in the paper) during  training. Likewise, EWC essentially doubles the size of the network during training, to create the Fisher information matrix used by this method. However, both our weight matrices and EWC's Fisher information matrix are discarded after training. So while the overall growth in the number of parameters is negligible, the number of parameters needed during training is surely higher than with vanilla SGD.}

 {\subsection{Choice of Hyperparameter}
We found a large $\lambda$ ($2 * 10^3$) for EWC constraint in Eqn.~\ref{Eqn:training_EWC_constrained}  can effectively prevent a large parameter drifting from old tasks.  However, if the $\lambda$ is too large, the strict constraint would hinder the learning of new tasks.}

 {We tested the hyperparameter $H$ for $M_t^i$ and $W_t^i$ from 1 to 2500. The more complex the task, the larger $H$ is needed for the BPN since it provides more degree of freedom to better learn beneficial perturabtions \citep{haeffele2017global,du2019gradient}. For example, The H is 25$\sim$255 for Permuted MNIST task and is 100 $\sim$ 900 for eight sequential object recognition tasks. However, if the H is too large (e.g, 2500 for eight object recognition tasks) that does not match the complexity of the task, the performance of the BPN would decrease.}

 {After 5 epochs of training, the network converged in Fig.~\ref{fig:quantative_results} as only 2 classes per task need to be trained in that figure (but see Tab.~\ref{tab:eighttasks} and Fig.~\ref{fig:100_permuted_MNIST_tasks} for the more complex 8-dataset where each task may have up to 200+ classes; this one required more epochs per task to converge, up to 300 epochs per dataset).}

 {\subsection{Difference between Transfer Learning and Continual Learning}}
 {Continual learning is a different idea from transfer learning.} 

 {For transfer learning, after learning the first task and when receiving a new task to learn, the parameters of the network are finetuned on the new task data. Thus, transfer learning is expected to suffer from forgetting the old tasks while being advantageous for the new task. Though, the shared convolutional layers benefits from a more general embeddings learned from a much more difficult task (pretrained on ImageNet model). }

 {In comparison, for continual learning, the focus is on learning new independent tasks sequentially without forgetting previous task. To achieve this focus, our BPN updates the shared normal weight using EWC or PSP. In theory, this update lead to orthogonal, hence non-overlapping and local task representation. However, in reality, the overlapping is inevitable because the parameters become too constrained for EWC, or PSP runs out of unrealized capacity of the core network. Thus, we introduce bias units trained independently for each task. At test time, bias units for a given task are activated to push representations of old tasks back to their initial task-optimal working regions.}

 {One of the benefits of continual learning is that learning new tasks can be aided by the knowledge already accumulated while learning the previous tasks.}

\subsection{Algorithms for BD + PSP}
The forward and backward rules for BD + PSP are detailed in Alg.~\ref{alg:FORTA_PSP} and Alg.~\ref{alg:BACTA_PSP}.

\setcounter{figure}{0}
\renewcommand{\thefigure}{S\arabic{figure}}%
\begin{figure*}[htb]
	\begin{center}
		\includegraphics[width=1\linewidth]{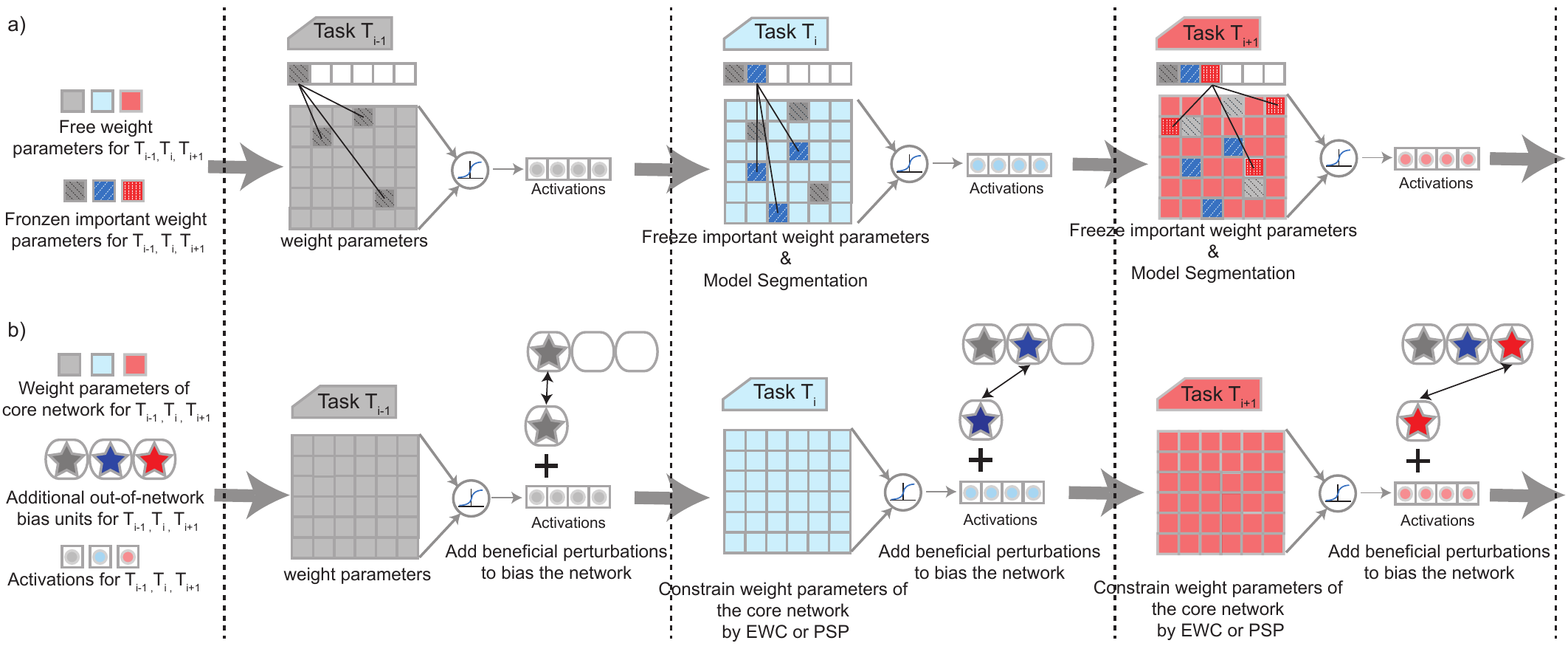}
		\caption{The dashed line indicates the start of a new task. (a) {\bf Flow chart of a typical Type 4 method}, here illustrated using a pictorial representation similar to that of Progressive Segmented Training (PST) \cite{du2019single}. PST subdivides the core network by freezing the important weight parameters for old tasks, and allowing new tasks to update the remaining free weight parameters. (b) {\bf Flow chart of Beneficial Perturbation Network (BPN; new type 5 method)}. BPN adds task-dependent beneficial perturbations to the activations, biasing the network toward that task, and retrains the weight parameters of the core network, with constraints from EWC \citep{kirkpatrick2017overcoming}, PSP\citep{cheung2019superposition}, or other similar approach. }
		\label{fig:Flow_charts}
	\end{center}
\end{figure*}

\setcounter{algorithm}{2}

\begin{figure*}[h!]
\begin{minipage}{\linewidth}

\begin{algorithm}[H]
\small
\caption{BD + PSP : forward rules for task t }
\label{alg:FORTA_PSP}

\begin{algorithmic}
\State {\quad For each fully connected layer i}
\State {\quad Select bias units (task t): $\mathbf{BIAS}_{t}^{i}$) for the current task }
\State{\quad{\bfseries Input:}\hspace{0.15cm} $\mathbf{BIAS}_{t}^{i}$ \textemdash
\hspace{0.08cm} Bias units for task t} \hfill {\color{green}// provide beneficial perturbations to bias the neural network}
\State{\quad\quad\quad\quad\hspace{0.33cm}$\mV^{i-1}$\textemdash
\hspace{0.08cm} Activations from the last layer}
\State{\quad\quad\quad\quad\hspace{0.33cm}$\mathbf{k}_{t}$\textemdash
\hspace{0.08cm} Binary keys for task t}
\State{\quad{\bfseries Output:} $\mV^{i} = \sigma(\mW^{i} \cdot  \mV^{i-1}  \odot \mathbf{k}_{t} + \mathbf{b}^{i} +\mathbf{BIAS}_{t}^{i}) \ \ \ \mathbf{\forall}  \ i \in [1,n]$ \hfill {\color{green}// activations for the next layer}}
\State{\quad\quad\quad\qquad \hspace{0.13cm} where:  $\mW^{i}$\textemdash
\hspace{0.08cm} normal neuron weights at layer $i$. $\mathbf{b}^{i}$\textemdash \hspace{0.08cm} normal bias term at layer $i$}
\State{\quad\quad\quad\qquad \hspace{1.13cm} $n$ \textemdash the number of layers \hspace{1.1cm} $\sigma(\cdot)$ \textemdash \hspace{0.08cm} the nonlinear activation function at each layer}
\end{algorithmic}
\end{algorithm}

\begin{algorithm}[H]
\small
\caption{BD + PSP : backward rules for task t }
\label{alg:BACTA_PSP}

\begin{algorithmic}

\State {{\underline {For the  task t:}}}
\State{{\quad Minimizing loss function: $L(\mX_{t},\mW^{i},\mathbf{BIAS}_{t}^{i})  \ \ \ \mathbf{\forall}  \ i \in [1,n]$}}
\State{{\quad \quad where: $\mX_{t}$\textemdash
\hspace{0.08cm} data for task One.\quad $\mW^{i}$\textemdash
\hspace{0.08cm} normal neuron weights at layer $i$. }}
\State{{  \quad\quad\quad\quad \hspace{0.16cm} $\mathbf{BIAS}_{t}^{i}$ \textemdash
\hspace{0.08cm}  bias units for task One from FC layers $i$, which is the product of $(\mM_{t}^{i}, \mW_{t}^{i})$}} 
\State{{\hspace{0.16cm}\quad\quad\quad\quad\quad $n$\textemdash
\hspace{0.08cm} is the number of FC layers.}}

\State{}

\State{\bf \underline{For each fully connected layer i:}}
\State{}
\State{\quad \underline {During the training of task t}}
\State {\quad \hspace{0.07cm} Select bias units for the current task t ($\mathbf{BIAS}_{t}^{i}$)}
\State{\hspace{0.07cm} \quad{\bfseries Input:}\hspace{0.15cm} $\mathbf{Grad}$ \textemdash
\hspace{0.08cm} Gradients from the next layer}
\State{\hspace{0.07cm} \quad{\bfseries output:} $\mathbf{dW_{t}^{i}} = \mathbf{Grad}\cdot((\mM_{t}^{i})^T)$  {\color{green}// gradients for the second term of bias units for task t at layer i } }
\State{\hspace{0.2cm}  \quad  $\hspace{33pt}$   $\mathbf{dM_{t}^{i}} = \epsilon\; sign\;((\mathbf{W}_{t}^{{i}})^T\cdot(\mathbf{Grad}))$ }
\State{\hspace{0.07cm} \hfill  {\color{green}// gradients for the first term of bias units for task t at layer i using FGSD method}}
\State{\hspace{0.07cm} \quad \hspace{0.07cm}  $\hspace{33pt}$   $\mathbf{dW^i} = \mathbf{Grad}\cdot((\mathbf{V}^{i})^T) \odot \mathbf{k}_{t}$\hfill {\color{green}// gradients for normal weights at layer i, $k_{t}$ is the key for task $t$}}
\State{\hspace{0.07cm} \quad \hspace{0.07cm}  $\hspace{33pt}$   $\mathbf{dV^{i}} = (\mathbf{W}^{i})^T \cdot (\mathbf{Grad}) \odot \mathbf{k}_{t}$ \hfill {\color{green}// gradients for activations at layer i to last layer i -1, $k_{t}$ is the key for task $t$}}
\State{\hspace{0.07cm} \quad  \hspace{0.07cm}  $\hspace{33pt}$   $\mathbf{db^i} =  \sum_{j} {Grad}_j $ \hfill {\color{green}// gradients for normal bias at layer i, j is iterator over the first dimension of {\bf{Grad}} }}
\State{}
\State {\quad \underline {After training of task t}}
\State{\quad \hspace{0.07cm}  Freeze the $\mathbf{BIAS}_{t}^{i}$}
\State{\quad \hspace{0.07cm}  Delete the $\mathbf{W_{t}^{i}}$ and  $\mathbf{M_{t}^{i}}$ to reduce parameter and memory storage cost}

\end{algorithmic}
\end{algorithm}

\end{minipage}
\end{figure*}

\end{document}